\documentclass[runningheads]{llncs}
\usepackage[final,year=2024,ID=1123]{accv}
\usepackage{accvabbrv}

\usepackage[T1]{fontenc}

\usepackage{graphicx}
\usepackage{booktabs}
\usepackage[accsupp]{axessibility}
\usepackage{amsmath}
\usepackage{amssymb}
\usepackage{mathtools}

\usepackage[breaklinks,colorlinks]{hyperref}
\usepackage{orcidlink}
\usepackage{multirow}
\usepackage{subcaption}

\graphicspath{ {./images/} }
\usepackage{booktabs}
\usepackage{multirow}
\usepackage{algorithm}
\usepackage{algorithmic}
\usepackage{enumitem}
\usepackage[flushleft]{threeparttable}
\usepackage{color}
\usepackage{diagbox}
\usepackage{array}

\def\ie{i.e.}
\def\eg{e.g.}
\def\st{\textrm{s.t.}}
\def\and{\textrm{and}}

\def\Diag{\textrm{Diag}}

\def\trace{\textrm{trace}}
\def\Diag{\textrm{Diag}}

\def\0{\textbf{0}}
\def\1{\textbf{1}}

\def\x{\boldsymbol{x}}

\def\z{\boldsymbol{z}}

\def\bX{\mathbf{X}}

\def\A{\mathcal{A}}
\def\A{\mathcal{A}}
\def\C{\mathcal{C}}

\def\L{\mathcal{L}}

\def\P{\mathcal{P}}

\def\X{\mathcal{X}}

\def\Z{\mathbf{Z}}
\def\RR{\mathbb{R}}

\newcommand{\myparagraph}[1]{\noindent\textbf{#1.}}

\usepackage[capitalize]{cleveref}
\crefname{section}{Sec.}{Secs.}
\Crefname{section}{Section}{Sections}
\Crefname{table}{Table}{Tables}
\crefname{table}{Tab.}{Tabs.}

\newcommand{\mcr}{\color{black}}
\newcommand{\mcb}{\color{black}}

\begin{document}

{\mcb 
\title{Graph Cut-guided Maximal Coding Rate Reduction for Learning Image Embedding and Clustering}

\titlerunning{Graph Cut-guided Maximal Coding Rate Reduction}

\author{Wei He\inst{1}\orcidID{0000-0003-2222-8818} \and
Zhiyuan Huang\orcidID{0009-0004-4047-3328}\inst{1} \and
Xianghan Meng\inst{1}\orcidID{0009-0006-4042-917X} \and
Xianbiao Qi\inst{2}\orcidID{0000-0002-8493-1966} \and
Rong Xiao\inst{3}\orcidID{0000-0003-2207-5698} \and
Chun-Guang Li$^{\ast,}$\inst{1}\orcidID{0000-0002-5716-268X}
}
\authorrunning{W. He, Z. Huang, X. Meng, X. Qi, R. Xiao, C.-G. Li}

\institute{Beijing University of Posts and Telecommunications,{~Beijing},{~P.R. China} \\
\email{\{wei.he, huangzhiyuan, mengxianghan, lichunguang\}@bupt.edu.cn} \\
\and
Visual Computing Group, International Digital Economy Academy (IDEA),{~Shenzhen},{~P.R. China} \and
Intellifusion,{~Shenzhen},{~P.R. China}
}

\maketitle
\begin{abstract}
In the era of pre-trained models, image clustering task is usually addressed by two relevant stages: a) to produce features from pre-trained vision models; and b) to find clusters from the pre-trained features. However, these two stages are often considered separately or learned by different paradigms, leading to suboptimal clustering performance. In this paper, we propose a unified framework, termed graph Cut-guided Maximal Coding Rate Reduction (CgMCR$^2$), for jointly learning the structured embeddings and the clustering. To be specific, we attempt to integrate an efficient clustering module into the principled framework for learning structured representation, in which the clustering module is used to provide partition information to guide the cluster-wise compression and the learned embeddings is aligned to desired geometric structures in turn to help for yielding more accurate partitions. We conduct extensive experiments on both standard and out-of-domain image datasets and experimental results validate the effectiveness of our approach.

\end{abstract}

\section{Introduction}
\label{sec:intro}

Image clustering, as a fundamental problem in computer vision and pattern recognition, aims to group images without annotated labels~\cite{Jain:1999-cluster}.
Image clustering is usually addressed by two successive stages: a) learning representation to generate features from images at first, and then b) finding clusters from the learned representation.
State-of-the-art methods for image clustering, \eg, subspace clustering algorithms~\cite{You:CVPR2016-EnSC,Chen:CVPR20,Lim:DSSC-arxiv20}, demonstrate remarkable performance on simple datasets such as MNIST~\cite{LeCun:PIEEE1998} and COIL~\cite{Nene:TR96} when proper features (\eg, Scatter Transform~\cite{Bruna:PAMI2013-scatter}) are provided.
In general, different clustering algorithms implicitly employ different assumptions about the geometry of the clusters. For instance, in $k$-means algorithm \cite{MacQueen-1967} each cluster is modeled as a standard Gaussian distribution and thus characterized by the mean vector;
in subspace clustering~\cite{Vidal:SPM11-SC}, each cluster is modeled as a low-dimensional subspace; in manifold clustering~\cite{Souvenir:ICCV2005-manifold} each cluster is modeled as a (low-dimensional) submanifold. 
Clustering performance will dramatically degenerate when the data distribution deviates from the assumption on the clusters. 

When dealing with challenging and complex datasets, such as CIFAR~\cite{Krizhevsky:CIFAR2009} and ImageNet~\cite{Deng:CVPR2009-imagenet}, the learned features play a more crucial role, as clustering based on conventional feature extraction typically fails to achieve satisfactory results.
The key ingredient among the recent remarkable progress in clustering is learning features by pre-trained models (\eg, auto-encoders~\cite{rumelhart:1986error, Kingma:ICLR2014-AE} and contrastive learning~\cite{Chen:ICML2020-contrastive}) that are suitable for the downstream clustering task.
In the era of pre-trained models, deep clustering methods are designed by learning features via visual pretraining and then learning the cluster membership from refined features~\cite{Van:ECCV2020-SCAN,Li:AAAI21-CC}.
More recently, when the large-scale pre-trained models, \eg, DINO~\cite{Caron:ICCV2021-DINO,Maxime:arXiv2023-DINOv2} and CLIP~\cite{Radford:ICML2021-CLIP}, are prevailing, deep clustering methods achieve impressive clustering performance by leveraging the rich representations produced by large-scale pre-trained models~\cite{{Adaloglou:BMVC2022-TEMI,Li:ICML24-TAC}}.
While promising clustering accuracy has been reported, these deep clustering methods are usually designed to learn the pseudo labels without leveraging the potential effect from the structures of embeddings.
There are a few attempts~\cite{Li:arXiv2022-NMCE,Ding:ICCV2023-MLC,Chu:ICLR2024-CPP} to learn the embeddings with desired structures---a \textit{union of orthogonal subspaces}, by leveraging the framework of Maximal Coding Rate Reduction (MCR$^2$)~\cite{Yu:NIPS20}. Nonetheless, none of these methods have developed a principled joint optimization framework to learn both the structured embeddings and the clustering.

In this paper, we present an effective joint optimizing framework, termed graph \textbf{C}ut-\textbf{g}uided \textbf{M}aximal \textbf{C}oding \textbf{R}ate \textbf{R}eduction (\textbf{CgMCR}$^2$), to learn both the structured embeddings and the clusters in principled way. 
Specifically, in CgMCR$^2$, we integrate both the normalized cut based clustering learning and the MCR$^2$ based structured representations learning to form a unified optimization problem.
Moreover, we design a two-stage training procedure, which %that 
consists of a one-shot initialization and a fine-tuning, to effectively train the proposed joint learning framework. 
We conduct extensive experiments on benchmark datasets to validate the superior performance of our proposed approach and also provide a set of ablation studies to evaluate the effect of each component. 
Our code is available at:~\url{https://github.com/hewei98/CgMCR2}.

\section{Relate Work}
\label{sec:relate}

{\mcr

\myparagraph{Pre-trained Vision models}
Typically, pre-trained models leverage self-supervised pretext tasks to learn representations from unlabeled datasets. For example, 
% AE
Autoencoders~\cite{Kingma:ICLR2014-AE} use an encoder-decoder architecture to learn latent low-dimensional representations by requiring the decoder to reconstruct the encoder inputs; 
contrastive learning (\ie, %the 
SimCLR~\cite{Chen:ICML2020-contrastive}) %emphasizes the importance of 
exploits the self-supervision information via data augmentation and learns representations that maximize the agreement between positive pairs and the disagreement between negative pairs;
and the following studies are proposed to improve the contrastive learning by enhancing the training stability (\eg, MoCo~\cite{He:CVPR2020-moco, chen:arxiv2020-mocov2}), reducing the requirement for negative samples (\eg, BYOL~\cite{Grill:NIPS2020-BYOL}), or avoiding the collapse solution of learned features (\eg, VICReg~\cite{Bardes:ICLR2022-vicreg}).
More recently, large-scale pre-trained models based on large models such as BERT~\cite{Devlin:2019-BERT} and Vision Transformers (ViTs)~\cite{Dosovitskiy:ICLR21-ViT} have showcased the capability to learn rich representations from diverse data sources.
For instance,
MAE~\cite{He:CVPR2022-MAE} leverages the ViT as the backbone of auto-encoder and uses the input images with a high mask proportion for training.
Inspired by self-supervised pretraining tasks in natural language processing, DINO~\cite{Caron:ICCV2021-DINO,Maxime:arXiv2023-DINOv2} implements a self-distillation framework with ViTs without using annotated labels. % supervision.
CLIP~\cite{Radford:ICML2021-CLIP} pretrains a vision-language model on image-text data pairs to learn visual concepts from a text-guided contrastive learning task.
%
% These large-scale pre-trained models have showcased their capability of learning rich representations from diverse data sources.
%
Although the pre-trained models mentioned above successfully learn meaningful features, the structure of the clusters in the learned features remains unclear.

\myparagraph{Clustering via pre-trained models}
The success of pre-trained models has led to breakthroughs in image clustering.
SCAN~\cite{Van:ECCV2020-SCAN} suggests a three-stage clustering pipeline: learning embeddings from a pre-trained model, using a clustering module for label prediction, and fine-tuning the clustering module using pseudo-labels;
%
%SCAN constructs its clustering algorithm based on the class-balanced assumption---assuming that all clusters have an equal number of data points.
%
RUC~\cite{Park:CVPR2021-Robust} and SPICE~\cite{Niu:IEEE2022-SPICE} enrich the pipeline of SCAN by utilizing robust training, refining network architecture 
or adjusting the fine-tuning strategy;
CC~\cite{Li:AAAI21-CC} and GCC~\cite{Zhong:ICCV21-GCC} propose unified frameworks for feature learning and clustering by optimizing the instance- and cluster-wise contrastive loss and graph contrastive loss, respectively.
IMC-SwAW~\cite{Ntelemis:2022-IMC} integrates a discrete representation into the self-supervised learning via a classifier net to simultaneously learn the cluster membership.
MiCE~\cite{Wei:ICLR21-MiCE} introduces a probabilistic clustering framework that combines contrastive learning with a latent mixture of experts.
ProPos~\cite{Huang:TPAMI22-ProPos} combines prototype scattering and positive sampling using EM-like steps to learn uniform, well-separated representations.
More recently, TEMI~\cite{Adaloglou:BMVC2022-TEMI} proposes a self-distillation clustering framework by leveraging large-scale pre-trained models;
and TAC~\cite{Li:ICML24-TAC} brings in the pre-trained CLIP texture embedding as external guidance for image clustering. 
Unfortunately, the intrinsic structure of representations learned or refined by the deep clustering methods mentioned above is still unclear, and thus usually only the nearest neighbors' information can be leveraged for clustering.
}

%\myparagraph{Clustering via the MCR$^2$ principle}
\myparagraph{Clustering via MCR$^2$}
%
%Given ground-truths, 
The %principle 
framework of Maximal Coding Rate Reduction (MCR$^2$)~\cite{Yu:NIPS20} is %able 
designed for supervised learning, to %to 
learn compact and structured representations %the instinct structure from images, %and produce within-cluster diverse and between-cluster discriminative embeddings.
that enjoy both the diversity in each class and the discriminativity between classes. 
There are some attempts to %utilize the principle 
use MCR$^2$ for deep clustering, \eg, 
NMCE~\cite{Li:arXiv2022-NMCE} designs a specialized self-supervised learning module %to produce initialize %representations
for initialization and subsequently optimizes the MCR$^2$ objective starting from the randomly initialized cluster membership;
MLC~\cite{Ding:ICCV2023-MLC} leverages the self-supervised learning module of NMCE for initialization %the embeddings learning 
and exploits a doubly stochastic affinity for partitioning the learned embeddings; 
then MLC is further evaluated on the CLIP pre-trained feature~\cite{Chu:ICLR2024-CPP}. 
While remarkable performance has been obtained, %there is still a lack of some principled and jointly trainable mechanism to find the clusters.
none of them have developed a principled joint framework to learn both the structured embeddings and the clustering.

\section{Our Approach: Graph Cut-guided Maximal Coding Rate Reduction (CgMCR$^2$)}
\label{sec:method}

We begin with a brief review of the principle of MCR$^2$ in Section \ref{subsec:preliminaries} and then present our framework for jointly learning structured embedding and clustering---CgMCR$^2$ in Section \ref{subsec:our_obj}.

\subsection{Preliminaries of MCR$^2$}
\label{subsec:preliminaries}

\myparagraph{Supervised feature learning via MCR$^2$}
Given a dataset $\X=\{\x_i\}_{i=1}^N$ of $N$ data points where $\x_i\in \RR ^D$ and the ground-truth labels $\mathbf \Pi^* \in \{0,1\}^{N \times k}$ to assign these $N$ points into $k$ 
classes $\C=\bigcup_{\ell=1}^k \C_\ell$.
The %principle 
framework of Maximal Coding Rate Reduction (MCR$^2$)~\cite{Yu:NIPS20} learns the embedding $\Z\in \RR^{d \times N}$ by %optimizing 
maximizing the following %rate reduction 
objective:
\begin{equation}
    \label{eq:mcr_supervised}
    \max_\Z \ R (\Z;\epsilon) - R_c (\Z, {\mathbf \Pi^*};\epsilon)
\end{equation}
where $R(\Z;\epsilon) := \log\det \left({\mathbf I} + \frac{d}{N\epsilon^2} \Z\Z^\top \right)$, $R_c (\Z, {\mathbf \Pi^*};\epsilon) :=\frac{1}{N}\sum_{\ell=1}^k N_\ell \log\det \big(\mathbf I + \\
\frac{d}{N_\ell\epsilon^2} \Z \Diag (\mathbf \Pi_\ell^*) \Z^\top \big)$,
and $N_\ell$ is the number of data points in class $\C_\ell$.
The first term $R (\Z;\epsilon)$ measures %the average coding length of the embeddings (a.k.a the coding rate) subject to a prescribed rate distortion precision $\epsilon > 0$, and 
the average coding length (a.k.a the coding rate) of the embeddings $\Z$ subject to a prescribed rate distortion precision $\epsilon > 0$, and 
%while 
the second term $R_c (\Z, {\mathbf \Pi^*};\epsilon)$ measures the sum of the coding rate of each 
class 
indicated 
by $\mathbf \Pi^*$.
%
%In short, 
Roughly, maximizing the second term $-R_c (\Z, {\mathbf \Pi^*};\epsilon)$ encourages the 
%
%
% within-cluster 
%
within-class 
embeddings to span a low-dimensional linear subspace; 
meanwhile, maximizing the first term $R (\Z;\epsilon)$ encourages %to expands 
the embeddings as a whole to expand, and thus 
making the class-specific subspaces 
being orthogonal to each other.
Such an arrangement of the class-specific subspaces is a desired % These properties of 
property of the embeddings learned by MCR$^2$ and %have 
has been proved in~\cite{Yu:NIPS20}. % for a given $\mathbf \Pi^*$.

\begin{figure*}[t]
    \centering
    \includegraphics[width=0.85\linewidth]{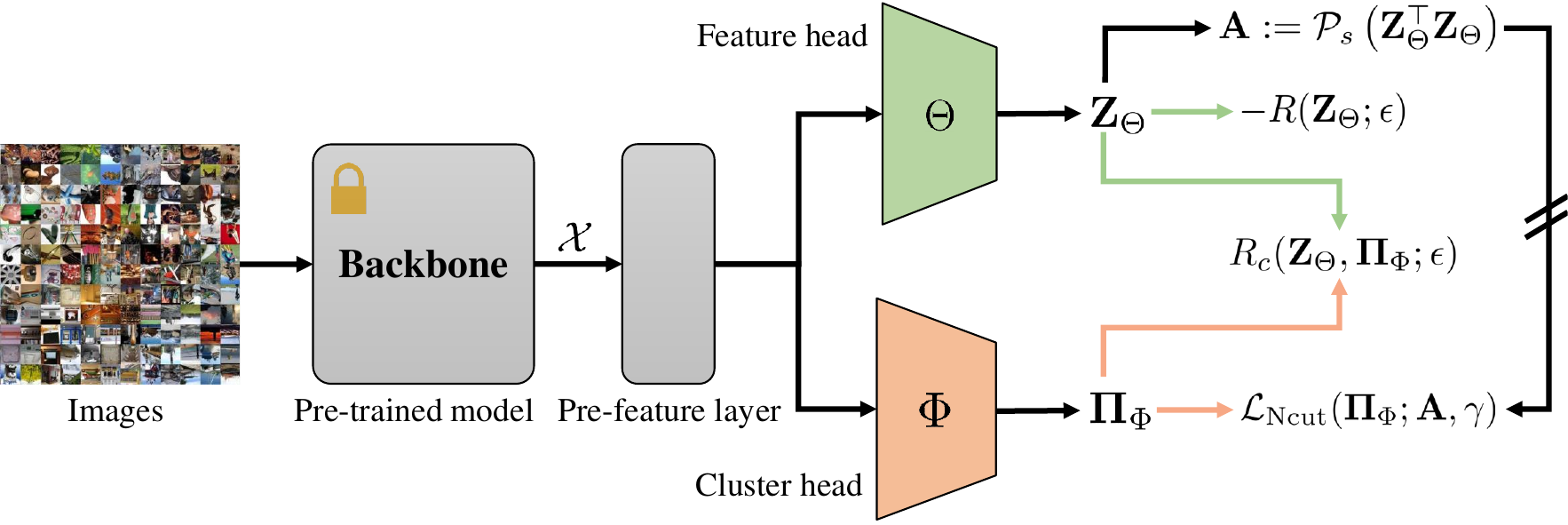}
  \caption{Illustration for our CgMCR$^2$ architecture. We illustrate the forward pass (in \textit{black} %lines
  ) and the gradient dependency of $\Z_\mathrm{\Theta}$ (in \textit{green} 
  % lines
  ) and $\mathbf{\Pi}_\mathrm{\Phi}$ (in \textit{orange} 
  %lines
  ) in different colored lines individually. For clarity, we exclude the parameters of the pre-feature layer from the visualization.}
  \label{fig:arch}
\end{figure*}

\myparagraph{Deep Clustering based on MCR$^2$}
The recent attempts~\cite{Li:arXiv2022-NMCE,Ding:ICCV2023-MLC,Chu:ICLR2024-CPP} employing the MCR$^2$ framework for deep clustering try to optimize the rate reduction objective to learn the representations and the partition %cluster memberships 
jointly as follows:
\begin{equation}
    \label{eq:mcr_unsupervised}
    \begin{split}
        \max_{\mathrm{\Theta}, \mathrm{\Phi}} \ R (\Z_\mathrm{\Theta};\epsilon) - R_c (\Z_\mathrm{\Theta}, \mathbf{\Pi}_\mathrm{\Phi};\epsilon), \quad \st ~~ \Z_\mathrm{\Theta}=f(\X;\mathrm{\Theta}), \mathbf{\Pi}_\mathrm{\Phi}=g(\X;\mathrm{\Phi}),
    \end{split}
\end{equation}
where $\Z_\mathrm{\Theta} \in \RR^{d\times N}$ and $\mathbf{\Pi}_\mathrm{\Phi} \in \RR^{N\times k}$ are implemented by feature head $f(\cdot;\mathrm{\Theta})$ and cluster head $g(\cdot;\mathrm{\Phi})$ which are parameterized by $\mathrm{\Theta}$ and $\mathrm{\Phi}$, respectively.
Although the representations $\Z_\mathrm{\Theta}$ and the partition $\mathbf{\Pi}_\mathrm{\Phi}$ are %cluster memberships 
jointly optimized, there lacks of an effective and principled mechanism to learn the partition (\ie, the cluster memberships) of the embeddings.

\subsection{Graph Cut-guided Maximal Coding Rate Reduction}
\label{subsec:our_obj}

Following the prior attempts based on MCR$^2$~\cite{Li:arXiv2022-NMCE,Ding:ICCV2023-MLC,Chu:ICLR2024-CPP} to learn both the embeddings and the partition, we try to incorporate a principled way to learn the cluster membership $\mathbf{\Pi}_\mathrm{\Phi}$ into the framework~\eqref{eq:mcr_unsupervised}.

\myparagraph{Spectral clustering for estimating $\mathbf{\Pi}_\mathrm{\Phi}$}
Spectral Clustering~\cite{vonLuxburg:StatComp07}, as a popular clustering algorithm with solid theoretical foundation, provides %offers 
a principled approach to learn the partition $\mathbf{\Pi}_\mathrm{\Phi}$ based on an affinity matrix built from the structured representation $\mathbf{Z}_\mathrm{\Theta}$.
Recall that, given an affinity 
%
%
% graph 
matrix $\mathbf{A}\in\RR^{N\times N}$ 
where entries $a_{i,j}$ measures the similarities of paired data points $(\x_i,\x_j)$, spectral clustering~\cite{vonLuxburg:StatComp07} aims to find an ideal cluster membership $\mathbf{\Pi}\in\{0,1\}^{N\times k}$ as follows: 
%
%which can be provably solved by optimizing following objective:
% 
%\begin{equation}
%label{eq:SC_origin}
%    \begin{split}
%    \min_{\tilde{\mathbf{\Pi}}} \ \trace \left(\tilde{\mathbf{\Pi}}^\top \mathbf{L} \tilde{\mathbf{\Pi}} \right),
%    \quad \st ~~~~ \tilde{\mathbf{\Pi}}\in\P
%    \end{split}
%\end{equation}
% 
\begin{equation}
\label{eq:SC_origin}
    \begin{split}
    \min_{{\mathbf{\Pi}}} \ \trace \left({\mathbf{\Pi}}^\top \mathbf{L} {\mathbf{\Pi}} \right),
    \quad \st ~~~~ {\mathbf{\Pi}}\in \{0,1\}^{N\times k} %\P
    \end{split}
\end{equation}
where $\mathbf{L}=\mathbf{D}-\mathbf{A}$ is known as the graph Laplacian corresponding to the affinity $\mathbf{A}$ and
$\mathbf{D} = \textrm{Diag}(d_{1,1},\cdots,d_{N,N})$ is the degree matrix with diagonal entries $d_{i,i}=\sum_{j=1}^N a_{i,j}$. 
% $\P$ is the set of feasible ${\tilde{\mathbf{\Pi}}}$, and $\tilde{\mathbf{\Pi}}$ is the normalized membership.
%
To avoid trivial partition, Normalized cut (Ncut)~\cite{Shi:IEEE2000-Ncut} introduces the volume of each partition to define the entries of $\mathbf{\Pi}$ as
%
%For instance, in Normalized cut (Ncut)~\cite{Shi:IEEE2000-Ncut}, the entries of $\tilde{\mathbf{\Pi}}$ are defined as
%
\begin{equation} 
\label{eq:Pi}
\tilde{\pi}_{i,\ell} :=
\begin{cases}
\frac{1}{\sqrt{|\mathcal{V}_\ell|}}& \ \text{if} \ \x_i \in \C_\ell\\
0& \ \text{if} \ \x_i \notin \C_\ell,
\end{cases}
\end{equation}
%
%
%where $\mathbf{V}_\ell:= \sum_{i:\x_i \in \C_\ell} d_{i,i}$ is the volume of $\ell$-th cluster.
%
%
where $|\mathcal{V}_\ell|:= \sum_{i:\x_i \in \C_\ell} d_{i,i}$ is the volume of $\ell$-th cluster. 
%
%Due to its combinatorial nature, 
%conventional spectral clustering approach solves a continuous relaxation of Eq.
%
%
%
Rather than solving the combinatorial problem in~\eqref{eq:SC_origin}, as a convention, spectral clustering methods \cite{vonLuxburg:StatComp07} reformulate the problem with continuous relaxation and then solve the relaxed problem. 
For example, Normalized cut (Ncut)~\cite{Shi:IEEE2000-Ncut} solves the problem as follows:
\begin{equation}
\label{eq:SC_relax}
    \begin{split}
    \min_{\tilde{\mathbf{\Pi}}} \ \trace \left(\tilde{\mathbf{\Pi}}^\top \mathbf{L} \tilde{\mathbf{\Pi}} \right),
    \quad \st ~~ \mathbf{\tilde \Pi}^\top \mathbf{D} \mathbf{\tilde \Pi} = \mathbf{I}.
    \end{split}
\end{equation}
%Solving this relaxed Ncut problem 
Note that %this is equivalently computing 
the solution is the ending $k$ eigenvectors associated with the ending $k$ minor eigenvalues of $\tilde{\mathbf L}=\mathbf{D}^{-\frac{1}{2}}\mathbf{L}\mathbf{D}^{-\frac{1}{2}}$.
%
%
%
%However, as the continuous relaxation in Eq.~\eqref{eq:SC_relax} gives up numerical constraints, the spectral embeddings produced by eigen-decomposition or contain negative elements and thereby can not sever as a direct probability membership.  
%
Unfortunately, the solution of the relaxed problem obtained by eigen-decomposition contains negative entries and can not be directly used as the cluster membership. Thus a $k$-means algorithm is adopted to generate the final clustering results. 
% In traditional spectral clustering, usually a $k$-means algorithm is applied on the spectral embeddings as a rounding heuristic step to obtain the cluster membership.

%
\myparagraph{Unified formulation of CgMCR$^2$}
Rather 
%
% 
% then 
%
%
than directly using the conventional NCut, %similar to~\cite{Anonymous:PR2024-neuncut}, 
we follow~\cite{He:PR2024-neuncut} % No longer need to anomynously cite it. To-do: list the authors and ... submitted to PR.
to relax the normalized cut problem as follows:
\begin{equation}
\begin{split}
\label{eq:neuncut}
%    \min_\mathbf{\Pi} \  \trace \left (\mathbf{\tilde \Pi}^\top \mathbf{L} \mathbf{\tilde \Pi} \right ) + \frac{\gamma}{2} \left \|\mathbf{\tilde \Pi}^\top \mathbf{D} \mathbf{\tilde \Pi} - \mathbf{I} \right \|_F^2, \quad \st ~~ 0\leq \mathbf{\Pi} \leq 1 , \ \mathbf{\Pi}\cdot\mathbf{1} = \mathbf{1}, 
%
     &\min_\mathbf{\Pi} \  \trace \left ((\mathbf{\Pi}\mathbf{V})^\top \mathbf{L} (\mathbf{\Pi} \mathbf{V}) \right ) + \frac{\gamma}{2} \left \| (\mathbf{\Pi}\mathbf{V})^\top \mathbf{D} (\mathbf{\Pi} \mathbf{V}) - \mathbf{I} \right \|_F^2, \\
\quad & \st ~~ 0\leq \mathbf{\Pi} \leq 1 , \ \mathbf{\Pi}\cdot\mathbf{1} = \mathbf{1}, 
\end{split}
\end{equation}
where $\gamma$ is a trade-off parameter, %the definitions of $\mathbf{L}$ and $\mathbf{D}$ are the same as that of in Eq.~\eqref{eq:SC_origin}, $\mathbf{\tilde \Pi} = \mathbf{\Pi} \mathbf{V}$, 
and $\mathbf{V}$ consists of estimated volumes for all the clusters, \ie:
\begin{equation}
\label{eq:vol_estimate}
    \mathbf{V} := \Diag\left( \sum_{i=1}^N \pi_{i, 1} d_{i,i}, \cdots , \sum_{i=1}^N \pi_{i, k} d_{i,i} \right)^{-1/2}.
\end{equation}
Note that the objective in Eq.~\eqref{eq:neuncut} enforces %emphasizes 
strict numerical constraints to have %of 
an ideal membership $\mathbf{\Pi}$ while relaxing the orthogonal constraints in Eq.~\eqref{eq:SC_relax} to a penalty term.
This relaxation %allows 
enable us to develop %for 
a %totally-differentiable 
differentiable approach for spectral clustering to learn the cluster membership directly (rather than the spectral embeddings). %More importantly, 
Interestingly, these strict numerical constraints %in Ncut problem 
can easily be %implicitly 
satisfied by reparameterizing $\mathbf{\Pi}$ with a neural network $\mathbf{\Pi}_\mathrm{\Phi}:=g(\cdot; \mathrm{\Phi})$ where $\mathrm{\Phi}$ denotes all the parameters in $g(\cdot)$ and a \texttt{softmax} function is used is used for the output.

By putting problems in \eqref{eq:mcr_unsupervised} and \eqref{eq:neuncut} together, %Consequently, 
we have an effective and principled joint optimization framework as follows:
\begin{equation}
\begin{split}
\label{eq:cgmcr2}
    \min_{\mathrm{\Theta}, \mathrm{\Phi}} \ \ -R(\Z_\mathrm{\Theta};\epsilon) +  R_c (\Z_\mathrm{\Theta}, \mathbf{\Pi}_\mathrm{\Phi};\epsilon) + \L_\mathrm{Ncut}(\mathbf{\Pi}_\mathrm{\Phi};\mathbf{A},\gamma),
\end{split}
\end{equation}
where $\L_\mathrm{Ncut}(\mathbf{\Pi}_\mathrm{\Phi};\mathbf{A},\gamma) = \trace \left (\mathbf{\tilde \Pi}_\mathrm{\Phi}^\top \mathbf{L} \mathbf{\tilde \Pi}_\mathrm{\Phi} \right )  + \frac{\gamma}{2} \left \|\mathbf{\tilde \Pi}_\mathrm{\Phi}^\top \mathbf{D} \mathbf{\tilde \Pi}_\mathrm{\Phi} - \mathbf{I} \right \|_F^2$, $\mathbf{\tilde \Pi}_\mathrm{\Phi}=\mathbf{\Pi}_\mathrm{\Phi} \mathbf{V}$. 
We use the cosine similarity of $\Z_\mathrm{\Theta}$ in current training iteration to %measure affinity graph, 
define the affinity, \ie, $\mathbf{A}:=\P_s\left(\Z_\mathrm{\Theta}^\top \Z_\mathrm{\Theta}\right)$, where $\P_s$ is a post-process operator. In practice, we simply reserve the $s$ largest entries of each row in $\mathbf{A}$. 
We termed this framework in~\eqref{eq:cgmcr2} as a graph \textbf{C}ut-\textbf{g}uided \textbf{M}aximal \textbf{C}oding \textbf{R}ate \textbf{R}eduction (\textbf{CgMCR}$^2$). 

\myparagraph{Remarks} 
The advantages of the unified CgMCR$^2$ %objective 
framework %in Eq.~\eqref{eq:cgmcr2} 
are \textit{three-folds}. 
\begin{itemize}
    \item %a) 
    The cluster membership is obtained by a principled way via $\L_\mathrm{Ncut}(\mathbf{\Pi}_\mathrm{\Phi};\mathbf{A},\gamma)$ and it can guide the optimization of $R_c (\Z_\mathrm{\Theta}, \mathbf{\Pi}_\mathrm{\Phi};\epsilon)$ for refining $\Z_\mathrm{\Theta}$. 
%Firstly, the cluster membership obtained by spectral clustering and the cluster membership required in the principle of MCR$^2$ are unified in Eq.~\eqref{eq:cgmcr2}. 
%In this objective, membership $\mathbf{\Pi}_\mathrm{\Phi}$ can be learned by optimizing $\L_\mathrm{Ncut}(\mathbf{\Pi}_\mathrm{\Phi};\mathbf{A},\gamma)$, which in turn, guides the optimizing of $R_c (\Z_\mathrm{\Theta}, \mathbf{\Pi}_\mathrm{\Phi};\epsilon)$ for refining $\Z_\mathrm{\Theta}$.
%
%Besides, the CgMCR$^2$ objective is completely differentiable, enabling the GPU acceleration for training neural networks.
\item 
%b) 
The framework is differentiable and scalable, enabling the GPU acceleration for training in mini-batch mode with stochastic gradient descent. % neural networks;
%Hence, our method can be scaled to large-scale datasets such as ImageNet, and can infer representations and cluster memberships for unseen data points in test sets.
\item 
%c) 
The framework can infer the representations and cluster memberships for unseen test data. % points in test sets.
\end{itemize}

\subsection{Implementations}
\label{subsec:implement}

\myparagraph{Network architecture}
For clarity, we illustrate our CgMCR$^2$ framework in Fig.~\ref{fig:arch}.
We utilize the frozen image encoder of CLIP~\cite{Radford:ICML2021-CLIP} %image encoder 
as a general feature extractor %to produce pre-features 
and %adopt 
add a single linear layer %as the pre-feature layer for scaling the pre-features' dimension.
to generate the pre-trained CLIP feature. 
We then deploy a two-layers neural network as the feature head $f(\cdot;\mathrm{\Theta})$ to learn the representation $\Z_\mathrm{\Theta}$, %that refines pre-features into target structures, 
and use a two-layers neural network with Gumbel-Softmax~\cite{Jang:ICLR2017-gumbel} %layer $g(\cdot;\mathrm{\Phi})$ 
as the cluster head $g(\cdot;\mathrm{\Phi})$ to learn the cluster membership $\mathbf{\Pi}_\mathrm{\Phi}$.

\myparagraph{Training procedure} 
%
%Still, optimizing the non-convex CgMCR$^2$ objective in Eq.~\eqref{eq:cgmcr2} poses challenges when representation $\Z_\mathrm{\Theta}$ and cluster membership $\mathbf{\Pi}_\mathrm{\Phi}$ are randomly distributed at the beginning of training.
%
%Therefore, 
% 
Taking into account the fact that optimizing $ R_c (\Z_\mathrm{\Theta}, \mathbf{\Pi}_\mathrm{\Phi};\epsilon)$ for learning $\Z_\mathrm{\Theta}$ could be %can be 
inefficient and challenging when $\mathbf{\Pi}_\mathrm{\Phi}$ is randomly initialized, %In this paper, 
we propose an efficient two-stage training strategy, which %containing 
consists of %an 
a \textit{one-shot initialization} stage and a \textit{fine-tuning} stage. 
% to learn the CgMCR$^2$ objective efficiently.

In the \textit{initialization} stage, we take %several warm-up 
$T_1$ epochs for warm-up to learn the initial discriminative representation and the initial %sub-optimal 
cluster membership by solving a %simpler 
simplified pretext task:
\begin{equation}
\begin{split}
\label{eq:pretext}
    \min_{\mathrm{\Theta}, \mathrm{\Phi}} \ \ -R(\Z_\mathrm{\Theta};\epsilon) + \L_\mathrm{Ncut}(\mathbf{\Pi}_\mathrm{\Phi};\mathbf{A},\gamma). 
\end{split}
\end{equation}
%
%where 
Note that the second term $ R_c (\Z_\mathrm{\Theta}, \mathbf{\Pi}_\mathrm{\Phi};\epsilon)$ in~\eqref{eq:cgmcr2} is temporally ignored due to the lack of a good initialization for $\mathbf{\Pi}_\mathrm{\Phi}$.   % removed.
%
% This is based on the fact that optimizing $ R_c (\Z_\mathrm{\Theta}, \mathbf{\Pi}_\mathrm{\Phi};\epsilon)$ for learning $\Z_\mathrm{\Theta}$ can be inefficient and challenging when $\mathbf{\Pi}_\mathrm{\Phi}$ is randomly initialized.
%
%{\mcr
%Note that what we want is to ensure that the learning of $\mathbf{\Z}_\mathrm{\Theta}$ will not be affected by the learning of $\mathbf{\Pi}_\mathrm{\Phi}$ before it has been well initialized.
%}
%
%
%
%
% Thus   ----> ??? any logical connection
%
% 
Moreover, we detach the gradients with respect to $\mathrm{\Theta}$ (due to the affinity $\mathbf{A}$) in the $\L_\mathrm{Ncut}(\mathbf{\Pi}_\mathrm{\Phi};\mathbf{A},\gamma)$ term from the back propagation and %there by 
thus the %parameters of 
feature head and 
the cluster head can be %are 
separately trained %learned by optimizing the term 
via the loss $-R(\Z_\mathrm{\Theta};\epsilon)$ and the loss $\L_\mathrm{Ncut}(\mathbf{\Pi}_\mathrm{\Phi};\mathbf{A},\gamma)$, respectively.

We then optimize the CgMCR$^2$ objective in Eq.~\eqref{eq:cgmcr2} for fine-tuning.
The well-initialized $\mathbf{\Pi}_\mathrm{\Phi}$ now %becomes 
provides a \textit{self-supervised information} %signal} 
that guides the learning of within-cluster compact representation %when optimizing 
via the term $R_c (\Z_\mathrm{\Theta}, \mathbf{\Pi}_\mathrm{\Phi};\epsilon)$.
%
%Due to the gradients detachment from $\bA$, %Hence, 
%optimizing the term $R_c (\Z_\mathrm{\Theta}, \mathbf{\Pi}_\mathrm{\Phi};\epsilon)$ with two learnable variables now becomes much easier.
%
Consequently, the feature head and the cluster head are jointly learned to %the 
perform within-cluster %diverse 
compact and between-cluster discriminative representation learning as well as the optimal cluster membership in the fine-tuning. 
The %entire 
whole procedure for training our CgMCR$^2$ is summarized in Algorithm~\ref{algo:overall}.

\begin{algorithm}[t]
	\caption{Procedure for graph Cut-guided Maximal Coding Rate Reduction}    %.}
	\label{algo:overall}
	\begin{algorithmic}[1]
		\STATE \textbf{Input:} $N$ images, $\epsilon,\gamma > 0$, warm-up epochs $T_1$, fine-tunning epochs $T_2$, post-processing operator $\P_s$ with sparsity $s$, batch size $n$, and learning rate $\eta$   %.
		\STATE \textbf{Initialization:} $t=0$, pre-feature $\X$, and all trainable network parameters $\mathrm{\Xi}$    % .
        \FOR {each $t =1, \cdots, T_1 + T_2$}
        %
        % {each $t \in \{1, \cdots, T_1 + T_2\}$}
        %
        %
            \STATE %Sample 
            Pick a mini-batch sample $\X^{(t)}$ with $n$ data points from $\X$
            \STATE \texttt{\# Forward pass}
            \STATE %Forward pass to 
            Compute $\Z^{(t)}_\mathrm{\Theta}$ and $\mathbf{\Pi}^{(t)}_\mathrm{\Phi}$
            \STATE Compute affinity $\mathbf{A}^{(t)}:=\P_s ( {\Z^{(t)\top}_\mathrm{\Theta}} {\Z^{(t)}_\mathrm{\Theta}} )$ and detach it from back propagation
            \STATE \texttt{\# Backward propagation}
            \IF{$t\leq T_1$}
            %% Back propagation
            \STATE Compute the gradients $\nabla_\mathrm{\Xi}$  with respect 
            to
            objective in~\eqref{eq:pretext}
            %% Back propagation
            \ELSE \STATE Compute the gradients $\nabla_\mathrm{\Xi}$  with respect %
            to
            objective in~\eqref{eq:cgmcr2}
            \ENDIF
            \STATE $\mathrm{\Xi}^{(t+1)} \leftarrow \mathrm{\Xi}^{(t)}  - \eta \nabla_\mathrm{\Xi}$
            \STATE $t \leftarrow t +1$
        \ENDFOR
		\STATE \textbf{Output: $\mathrm{\Xi}^{(t+1)}$ }
		\end{algorithmic}
\end{algorithm}

\myparagraph{Comparison to %with 
NMCE, MLC and CPP}
In Table~\ref{tab:compare_mcr2}, we summarize the connections and differences between our CgMCR$^2$ and other deep clustering methods based on the framework of MCR$^2$.
These works proceed by initializing representations through pre-trained models, and refining the representation and cluster membership after the initialization, in which 
NMCE~\cite{Li:arXiv2022-NMCE} and MLC~\cite{Ding:ICCV2023-MLC} leverage a specially designed objective that incorporates the contrastive learning objective ($\L_\mathrm{SimCLR}$)~\cite{Chen:ICML2020-contrastive} for pretraining,
%
%W
whileas CPP~\cite{Chu:ICLR2024-CPP} utilizes the pre-trained CLIP features. % to produce pre-feautres.
Among these works, $\mathbf{\Pi}_\mathrm{\Phi}$ is either randomly initialized (as in NMCE) or initialized by copying the parameters $\mathrm{\Theta}$ to $\mathrm{\Phi}$ (as in MLC and CPP). 
However, none of these methods employ a principled approach %method for initializing the cluster head, 
to learn the cluster membership in their optimization framework. 
% which is crucial for optimizing the non-convex objective as in Eq.~\eqref{eq:mcr_unsupervised}, has not been proposed.
%
%
As a result, NMCE %can be 
is sensitive to the random initialization, whereas MLC and CPP rely on additional clustering algorithms to obtain cluster memberships.
%
%By contrast, our framework CgMCR$^2$ %  objective and offers
%jointly learns both the structured representation and the cluster membership, in particular the cluster membership is learned in a principled way. 

%a principled training strategy to initialize both representation and cluster membership to guide the refining of these two variables.

\begin{table}[t]
\centering
\caption{\textbf{Comparing our CgMCR$^2$ to deep clustering methods via MCR$^2$.}}
\setlength{\tabcolsep}{0.8mm}{
    \begin{tabular}{l cccc}
    \toprule
    Methods & Off-the-shelf pretrain & Initializing $\Z_\mathrm{\Theta}$ & Initializing $\mathbf{\Pi}_\mathrm{\Phi}$ & Objective \\
    \midrule
    NMCE & $\times$ & $\mathrm{\Theta}\leftarrow \nabla_\mathrm{\Theta} (R+\L_\mathrm{SimCLR})$ & Random & MCR$^2$ \\
    MLC & $\times$ & $\mathrm{\Theta}\leftarrow \nabla_\mathrm{\Theta} (R+\L_\mathrm{SimCLR})$ & $\mathrm{\Phi} \leftarrow \mathrm{\Theta}$ & MCR$^2$ \\
    CPP & $\checkmark$ & $\mathrm{\Theta}\leftarrow \nabla_\mathrm{\Theta} R$ & $\mathrm{\Phi} \leftarrow \mathrm{\Theta}$ & MCR$^2$ \\
    % \midrule
    \textbf{CgMCR$^2$} & $\checkmark$ & $\mathrm{\Theta}\leftarrow \nabla_\mathrm{\Theta}R$ & $\mathrm{\Phi} \leftarrow \nabla_\mathrm{\Phi} \L_\mathrm{Ncut}$ & CgMCR$^2$ \\
    \bottomrule
    \end{tabular}}
\label{tab:compare_mcr2}
\end{table}

\section{Experimental Results}
\label{sec:result}
\myparagraph{Datasets}
We evaluate the performance of our CgMCR$^2$ on MNIST~\cite{LeCun:PIEEE1998}, Fashion MNIST (F-MNIST)~\cite{Xiao:FashionMNIST19}, CIFAR-10, CIFAR-20, CIFAR-100~\cite{Krizhevsky:CIFAR2009}, Stanford Dogs-120 (Dogs-120)~\cite{Khosla:CVPR2011-dogs}, Oxford Flowers-102 (Flowers-102)~\cite{Data:Flowers}, TinyImageNet-200 (TinyImageNet) and ImageNet-1k.
% These datasets are detailed in the supplementary.
Among these datasets, CIFAR-20 contains the same data as CIFAR-100 with 20 super-classes as %can be found 
in \cite{Krizhevsky:CIFAR2009}, whileas Stanford Dogs-120, TinyImageNet and ImageNet-1k are subsets of ImageNet~\cite{Deng:CVPR2009-imagenet}.
All data points are embedded to 768-dimensional vectors using a CLIP~\footnote{\url{https://github.com/openai/CLIP}} 
image encoder pre-trained on the ViT-L/14 backbone, or embedded to 512-dimensional vectors using the  MoCov2~\footnote{\url{https://github.com/facebookresearch/moco}} 
model pre-trained on the ResNet-18 backbone.

\myparagraph{Metrics}
To evaluate the clustering performancce, we report the clustering accuracy (ACC) and the normalized mutual information (NMI). For ACC, we use the Hungarian matching algorithm~\cite{kuhn:1955-hungarian} to find the best match between the pseudo-labels and ground-truth labels. By default, the pseudo-labels of CgMCR$^2$ are generated by the cluster head, \ie, the output of the $\texttt{argmax}$ layer. 

\myparagraph{Training settings}
We use Adam optimizer \cite{Kingma:ICLR2014} with a fixed initial learning rate during the warm-up, and then using cosine annealing learning rate \cite{Loshchilov:ICLR17} during the fine-tuning.
For all datasets, the output dimension $k$ of cluster head is set to the number of true clusters.
% All other hyper-parameters for different datasets are shown in the supplementary.

\subsection{Experiments %Clustering 
on Standard Datasets}

% \myparagraph{CgMCR$^2$ is scalable feature learner and label predictor}
\myparagraph{Visualization of Affinity Matrices}
To demonstrate the ability of CgMCR$^2$ to produce both structured features and satisfactory clustering results during the training process, 
%For that purpose, 
we conduct a set of experiments on CIFAR-10 to visualize the %ground-truth similarity 
affinity matrices of the original CLIP features $\X$, the feature head outputs $\Z_\mathrm{\Theta}$ and cluster head outputs $\mathbf{\Pi}_\mathrm{\Phi}$. Visualization results are given as colored images in Fig.~\ref{fig:visual_aff}.
\begin{figure}[t]
    \centering  % $|\X^\top \X|$
    \includegraphics[width=0.95\linewidth]{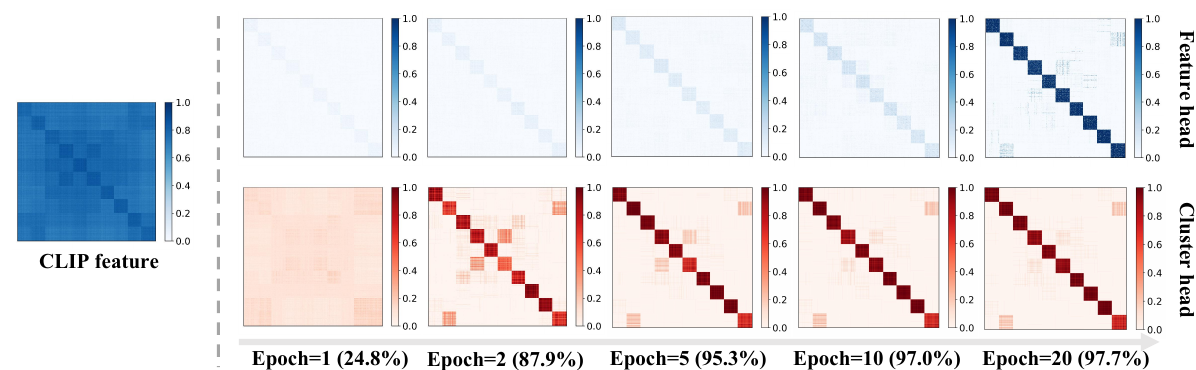}
  \caption{\textbf{Similarity matrices ordered by the ground-truth labels} of the CLIP features computed by $|\bX^\top \bX|$ (in the \textit{left panel in blue}), representation computed by $|\mathbf{Z}_\mathrm{\Theta}^\top \mathbf{Z}_\mathrm{\Theta}|$ (at the first row \textit{in blue}) and cluster membership $|\mathbf{\Pi}_\mathrm{\Phi} \mathbf{\Pi}_\mathrm{\Phi}^\top|$ (at the second row \textit{in red}) of CgMCR$^2$ trained with $\{1,2,5,10,20\}$ epochs on CIFAR-10, where the percentage number in bracket is ACC.}
  \label{fig:visual_aff}
\end{figure}
As can be seen that, our framework learns %between-cluster discriminative 
structured representations (see, \eg, the block diagonal %\textit{off-diagonal} blocks 
structure in \textit{blue} at epoch 10) and produces %sub-optimal 
acceptable initial cluster membership via the pretext task in~\eqref{eq:pretext}.
Then, fine-tuning the framework using the whole CgMCR$^2$ objective in~\eqref{eq:cgmcr2} refines the structured representations (see, \eg, the block \textit{diagonal} structure %blocks 
in \textit{blue} at epoch 20).
These structured representations, in turn, enable the cluster head to learn better cluster membership.
Once trained, the cluster head can serve as a scalable predictor to produce satisfactory pseudo labels, achieving 97.7\% accuracy on CIFAR-10.
\myparagraph{Performance Comparison between Feature Head and Cluster Head}
%We also demonstrate the performance of the feature head and cluster head.
For a trained CgMCR$^2$, rather than using the clustering head to yield the pseudo labels, we can also apply the conventional Spectral Clustering \cite{Shi:IEEE2000-Ncut} to the affinity which is defined by $\mathbf{A}:=\P_s\left(\Z_\mathrm{\Theta}^\top \Z_\mathrm{\Theta}\right)$. This method is denoted as ``SC on  $\Z_\mathrm{\Theta}$''. %As shown 
We report the comparison results in Table~\ref{tab:head_vs_head}. As can be read, both the feature head and the cluster head perform almost equally good. %can function as a structured feature extractor and a scalable pseudo-label predictor, respectively.
In the following experiments, we also denote the performance of ``SC on  $\Z_\mathrm{\Theta}$'' as ``\textbf{CgMCR$^2$-SC}'' to distinguish it from the performance obtained by ``$\texttt{argmax}$ on $\mathbf{\Pi}_\mathrm{\Phi}$'' (\textbf{CgMCR$^2$}) for clearity.

\begin{table*}[t]
\centering
\caption{\textbf{Clustering accuracy} (mean$\pm$std) of feature head (\textit{Top}) and the cluster head (\textit{Bottom}) using CLIP features on five benchmark datasets over 3 trials.}
\setlength{\tabcolsep}{1mm}{
    \begin{tabular}{lccccccccccc}
    \toprule
    Methods & CIFAR-10 & CIFAR-20 & CIFAR-100 & TinyImageNet & ImageNet-1k \\
    \midrule
    SC on $\Z_\mathrm{\Theta}$ & 97.6$\pm$0.1 & 68.8$\pm$0.4 & 78.3$\pm$0.3 & 72.7$\pm$0.2 & 67.7$\pm$0.2 \\
    $\texttt{argmax}$ on $\mathbf{\Pi}_\mathrm{\Phi}$ & 97.7$\pm$0.1 & 68.1$\pm$0.4 & 77.8$\pm$0.4 & 72.9$\pm$0.2 & 67.5$\pm$0.3 \\
    \bottomrule
    \end{tabular}}
\label{tab:head_vs_head}
\end{table*}

\myparagraph{Comparison %with 
to Competing Clustering Methods using CLIP Features}
To evaluate the performance of our CgMCR$^2$, we conduct experiments on five datasets and compare to a set of competing baseline methods.
As the baseline, we choose the classical clustering algorithms, including $k$-means~\cite{MacQueen-1967} and spectral clustering with normalized cut~\cite{Shi:IEEE2000-Ncut}, %~\cite{vonLuxburg:StatComp07} and 
subspace clustering method, Elastic Net Subspace Clustering (EnSC)~\cite{You:CVPR2016-EnSC}, deep clustering methods, including SCAN~\cite{Van:ECCV2020-SCAN}, TEMI~\cite{Adaloglou:BMVC2022-TEMI}, CPP~\cite{Chu:ICLR2024-CPP} and TAC~\cite{Li:ICML24-TAC}. 
All these methods are conducted on the CLIP features. We report the experimental results in Table~\ref{tab:compare_clip}. 
The results of CPP and TAC are reproduced with the released codes. The results of TEMI~\cite{Adaloglou:BMVC2022-TEMI} are cited from the paper.
We can read that, all methods yield promising performance owning to the CLIP feature. But, clearly, our CgMCR$^2$ achieves superior clustering performance.
The performance improvements over CPP~\cite{Chu:ICLR2024-CPP}, which is also based on the framework of MCR$^2$, are due to the principled way to produce the clustering membership. 
We note that TAC~\cite{Li:ICML24-TAC} also yields competitive results, but it leverages the external information brought by the CLIP text encoder.

\begin{table}[t]
\centering
\caption{\textbf{Clustering Performance Comparison using CLIP Features on Five Benchmark Datasets.} %to competing clustering methods and our CgMCR$^2$ using CLIP features on 5 benchmark datasets. 
`-' denotes that %not reported 
the results are not available.}
\setlength{\tabcolsep}{1mm}{
    \begin{tabular}{l   cc   cc   cc   cc   cc}
    \toprule
    \multirow{2}{*}{Methods} & \multicolumn{2}{c}{CIFAR-10} & \multicolumn{2}{c}{CIFAR-20} & \multicolumn{2}{c}{CIFAR-100} & \multicolumn{2}{c}{TinyImgNet} & \multicolumn{2}{c}{ImageNet-1k}\\
    & ACC & NMI & ACC & NMI & ACC & NMI & ACC & NMI & ACC & NMI \\
    \midrule
    $k$-means~\cite{MacQueen-1967} & 83.5 & 84.1 & 46.9 & 49.4 & 52.8 & 66.8 & 54.1 & 73.4 & 53.9 & 79.8 \\
    Spectral~\cite{Shi:IEEE2000-Ncut} & 79.8 & 84.8 & 53.3 & 61.6 & 66.4 & 77.0 & 62.8 & 77.0 & 56.0 & 81.2 \\
    EnSC~\cite{You:CVPR2016-EnSC} & 95.4 & 90.3 & 61.0 & 68.7 & 67.0 & 77.1 & {64.5} & {77.7} & 59.7 & {83.7} \\
    %SCAN$_\mathrm{CLIP}$~\cite{Van:ECCV2020-SCAN} & 95.1 & 90.3 & 60.8 & 61.8 & 64.1 & 70.8 & 56.5 & 72.7 & 54.4 & 76.8 \\
    SCAN~\cite{Van:ECCV2020-SCAN} & 95.1 & 90.3 & 60.8 & 61.8 & 64.1 & 70.8 & 56.5 & 72.7 & 54.4 & 76.8 \\
    TEMI~\cite{Adaloglou:BMVC2022-TEMI} & 96.9 & 92.6 & 61.8 & 64.5 & 73.7 & 79.9 & - & - & {64.0} & - \\
    CPP~\cite{Chu:ICLR2024-CPP} & {97.4} & {93.6} & {64.2} & {72.5} & {74.0} & {81.8} & 63.4 & 77.3 & 62.0 & 82.1 \\
    % \midrule
    {\mcb \textbf{CgMCR$^2$} } & \bf{97.7} & \bf{94.3} & \underline{68.1} & \underline{73.8} & \underline{77.8} & \underline{81.9} & \bf 72.9 & \bf 81.4 & \underline{67.5} & \underline{87.0} \\
    {\mcb \textbf{CgMCR$^2$-SC} } & \underline{97.6} & \underline{94.2} & \bf 68.8 & \bf 74.0 & \bf 78.3 & \bf 82.5 & \underline{72.7} & \underline{81.1} & \textbf{67.7} & \textbf{87.1} \\
    % \textbf{CgMCR$^2$} & \bf{97.7} & \bf{94.3} & \bf 68.8 & \bf 74.0 & \bf 78.3 & \bf 82.5 & \bf 72.9 & \bf 81.4  & \textbf{67.7} & \textbf{87.1} \\
    \midrule
    \multicolumn{11}{l}{\textit{External texture guidance}} \\
    {\mcb TAC~\cite{Li:ICML24-TAC} } & 97.0 & 92.4 & 66.8 & 73.2 & 75.5 & 81.1 & 71.0 & 79.9 & 66.4 & 86.8 \\
    \bottomrule
    \end{tabular}}
\label{tab:compare_clip}
\end{table}

\myparagraph{Comparison to State-of-the-art Deep Clustering Methods using MoCov2 Features}
We apply our CgMCR$^2$ framework on the pre-trained MoCov2~\cite{chen:arxiv2020-mocov2} features, and compare %it with
to %classical clustering algorithms and competing 
state-of-the-art deep clustering methods, including CC~\cite{Li:AAAI21-CC}, GCC~\cite{Zhong:ICCV21-GCC}, SCAN~\cite{Van:ECCV2020-SCAN}, SPICE~\cite{Niu:IEEE2022-SPICE}, IMC-SwAV~\cite{Ntelemis:2022-IMC}, NMCE~\cite{Li:arXiv2022-NMCE} and MLC~\cite{Ding:ICCV2023-MLC}. Also, we report the performance of $k$-means~\cite{MacQueen-1967}, Spectral Clustering 
%with normalized cut
\cite{Shi:IEEE2000-Ncut} and Elastic Net Subspace Clustering (EnSC)~\cite{You:CVPR2016-EnSC}. 
All methods, %in Table~\ref{tab:compare_contrast}, 
except for NMCE, MLC and IMC-SwAV, are conducted on the pre-trained MoCov2 features; whereas NMCE, MLC and IMC-SwAV are performed with %employ 
their specially designed pre-trained models, respectively. 
The results of TEMI~\cite{Adaloglou:BMVC2022-TEMI} is reproduced by using the released codes on the MoCov2 features.
For a relatively fair comparison, all method use ResNet18 as the backbone of the pre-trained models.
Experimental results are listed in Table~\ref{tab:compare_contrast}. %using pre-trained MoCov2 features. 
As can be read that, our CgMCR$^2$ still achieves %superior
the leading clustering accuracy on majority cases.

\begin{table}[t]
\centering
\caption{\textbf{Clustering Performance Comparison to State-of-the-art Deep Clustering Methods.}
}
\setlength{\tabcolsep}{2mm}{
    \begin{tabular}{l   cc   cc   cc   cc}
    \toprule
    \multirow{2}{*}{Methods} & \multicolumn{2}{c }{CIFAR-10} & \multicolumn{2}{c }{CIFAR-20} & \multicolumn{2}{c }{CIFAR-100} & \multicolumn{2}{c}{TinyImageNet} \\
    & ACC & NMI & ACC & NMI & ACC & NMI & ACC & NMI \\
    \midrule
    $k$-means~\cite{MacQueen-1967} & 71.6 & 61.8 & 41.4 & 41.3 & 42.8 & 42.6 & 17.8 & 41.3 \\
    Spectral~\cite{Shi:IEEE2000-Ncut} & 80.7 & 69.2 & 45.4 & 42.6 & 40.9 & 56.9 & 20.3 & 36.8 \\
    EnSC~\cite{You:CVPR2016-EnSC} & 83.2 & 74.1 & 50.6 & 45.5 & 41.2 & 62.7 & 25.3 & 40.4 \\
    {\mcb  CC~\cite{Li:AAAI21-CC} } & 79.0 & 70.5 & 42.9 & 43.1 & 36.9 & 58.1 & 24.0 & 44.0\\
    {\mcb  GCC~\cite{Zhong:ICCV21-GCC} } & 85.6 & 76.4 & 47.2 & 47.2 & 38.2 & 59.9 & 23.8 & 44.7 \\
    SCAN~\cite{Van:ECCV2020-SCAN} & 88.3 & 79.7 & 50.7 & 48.6 & 34.3 & 55.7 & - & - \\
    SPICE~\cite{Niu:IEEE2022-SPICE} & {91.8} & {85.0} & {53.5} &  \bf 56.5 & - & - & 30.5 & 44.9 \\
    IMC-SwAV~\cite{Ntelemis:2022-IMC} & 89.7 & 81.8 & 51.9 & 52.7 & 45.1 & 67.5 & 27.9 & 48.5 \\
    %    {\mcb TEMI$_\mathrm{MoCov2}$~\cite{Adaloglou:BMVC2022-TEMI} } & 88.7 & 84.1 & 47.3 & 48.7 & 45.4 & 66.7 & 27.8 & 42.3 \\    
    {\mcb TEMI~\cite{Adaloglou:BMVC2022-TEMI}  } & 88.7 & 84.1 & 47.3 & 48.7 & 45.4 & 66.7 & 27.8 & 42.3 \\
    NMCE~\cite{Li:arXiv2022-NMCE} & 83.0 & 76.1 & 43.7 & 48.8 & 40.0 & 53.9 & 21.6 & 40.0 \\
    MLC~\cite{Ding:ICCV2023-MLC} & 86.3 & 78.3 & 52.2 & \underline{54.6} & {49.4} & \bf 68.3 & {33.5} & \bf 67.5 \\
    % \midrule
    {\mcb \textbf{CgMCR$^2$} } & \bf 92.8 & \bf 88.6 & \underline{55.6} & 54.3 & \underline{49.8} & 67.9 & \underline{35.9} & 62.5\\
    {\mcb \textbf{CgMCR$^2$-SC} } & \underline{92.7} & \underline{88.4} & \bf 56.1 & 54.4 & \bf 51.1 & \underline{68.0} & \bf 36.7 & \underline{62.9} \\
    % \textbf{CgMCR$^2$} & \bf 92.7 & \bf 88.6 & {\bf 56.1} & 54.4 & {\bf 51.1} & \underline{68.0} & {\bf 35.9} & \underline{62.9} \\
    \bottomrule
    \end{tabular}}
\label{tab:compare_contrast}
\end{table}

%\subsection{Clustering on Out-of-Domain Datasets}
\subsection{Experiments on Out-of-Domain Datasets} % ?? 

To demonstrate the effectiveness of jointly learning both the structured representation and the clustering, we apply the pre-trained CLIP~\cite{Radford:ICML2021-CLIP} to extract features for datasets MNIST, F-MNIST, Flowers-102 and Dogs-120, which are quite different from the training data for the pre-trained CLIP. 
%
% In Table~\ref{tab:extra_dataset}, 
We %conduct experiments to 
compare the performance of our CgMCR$^2$ %with 
to four representative clustering methods, including %where 
Spectral Clustering, EnSC, SCAN and CPP. %and SCAN use fixed CLIP features, and CPP uses CLIP as pre-features.
Experimental results are listed in Table~\ref{tab:extra_dataset}. 
We can observe that, our CgMCR$^2$ %surpasses the limits of less satisfied CLIP features, achieving 
still yields satisfactory clustering accuracy on out-of-domain datasets, especially with a notable accuracy improvement of +10.3\% on Dogs-120.
%Additionally, 
Besides, we notice of that SCAN fails on Flowers-102, which is an \textit{imbalanced} dataset. This is because that SCAN is designed on a class-balance assumption, which is unsatisfied on Flowers-102. On contrary, our CgMCR$^2$ still produces satisfactory clustering results.

%Flowers-102 is \textit{imbalanced}, with each class containing 40 to 258 images.
%While clustering methods built on the assumption of class-balance, such as SCAN, fail on imbalanced datasets, our CgMCR$^2$ still produces satisfactory clustering results.
%

%
\begin{table}[th]
\centering
\caption{\textbf{Clustering Performance on Out-of-domain Datasets.} 
%We compare our CgMCR$^2$ to four representative clustering methods %the proposed CgMCR$^2$ on CLIP features.
}
\setlength{\tabcolsep}{2mm}{
    \begin{tabular}{l  cc  cc  cc  cc}
    \toprule
    \multirow{2}{*}{Methods} & \multicolumn{2}{c}{MNIST} & \multicolumn{2}{c}{F-MNIST} & \multicolumn{2}{c}{Flowers-102} &\multicolumn{2}{c}{Dogs-120} \\
    & ACC & NMI & ACC & NMI & ACC & NMI & ACC & NMI \\
    \midrule
    % \textit{Fixed CLIP} \\
    Spectral~\cite{Shi:IEEE2000-Ncut} & 74.5 & 67.0 & 64.3 & 56.8 & 85.6 & 94.6 & 44.1 & 55.6  \\
    EnSC~\cite{You:CVPR2016-EnSC} & 91.0 & 85.3 & 69.1 & 65.1 & 90.0 & 95.9 & 40.1 & 60.8 \\
    SCAN~\cite{Van:ECCV2020-SCAN} & 87.4 & 81.9 & 69.3 & 67.2 & 40.4 & 68.7 & 38.1 & \underline{73.8} \\
    CPP~\cite{Chu:ICLR2024-CPP} & {95.7} & {90.4} & {70.9} & {68.8} & \underline{91.3} & \underline{96.4} & {51.0} & 69.5 \\
    % \midrule
    % \textit{Refined CLIP} \\
    {\mcb \textbf{CgMCR$^2$} } & \bf 96.9 & \bf 92.8 & \underline{74.5} & \underline{69.9} & 91.1 & 96.1 & \underline{60.9} & {73.6} \\
    {\mcb \textbf{CgMCR$^2$-SC} } & \underline{96.4} & \underline{92.0} & \bf 74.7 & \bf 71.1 & \bf 92.2 & \bf 97.0 & \bf 61.3 & \bf 75.1 \\
    % \textbf{CgMCR$^2$} & \bf{96.4} & \bf{92.0} & \bf 74.7 & \bf 71.1 & \bf 92.2 & \bf 97.0 & \bf 61.3 & \bf 75.1 \\
    \bottomrule
    \end{tabular}}
\label{tab:extra_dataset}
\end{table}

\subsection{More Evaluation and Analysis}

\begin{figure}[t]
    \centering
    \begin{subfigure}[t]{0.45\textwidth}
        \includegraphics[width=\linewidth]{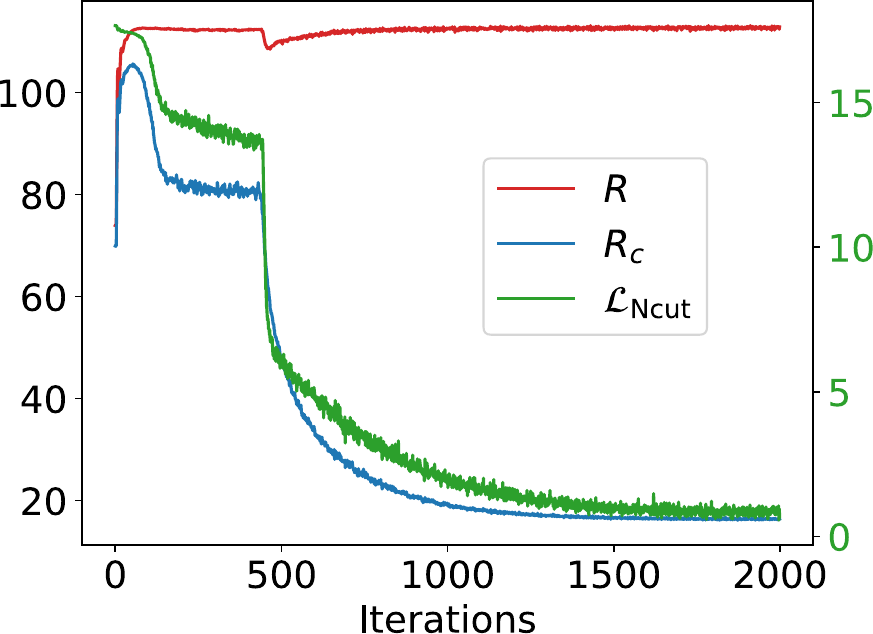}
        \caption*{(a)}
        %\caption{\textbf{Learning curves} of each term in Eq.~\eqref{eq:cgmcr2} on CIFAR-20.The range of $\L_\mathrm{Ncut}$ is displayed with the green-colored axis.}
    %\label{fig:obj_curve}
    \end{subfigure}
     \begin{subfigure}[t]{0.45\textwidth}
     \includegraphics[width=\linewidth]{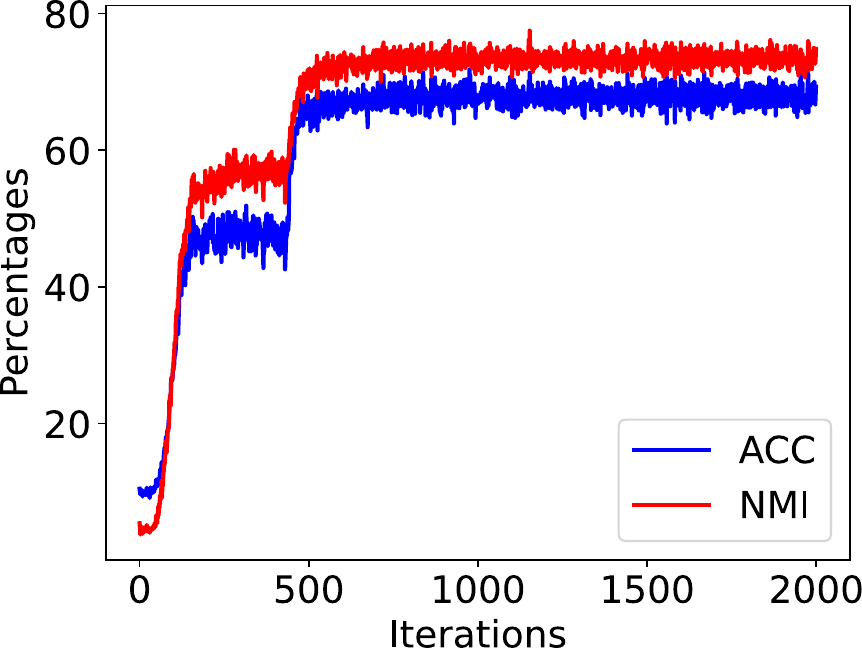}
     %\caption{\textbf{ACC and NMI curves} of the proposed CgMCR$^2$ on CIFAR-20.}
     \caption*{(b)}
    %\label{fig:acc_curve}
     \end{subfigure}%\hfill
     %\begin{minipage}[t]{0.32\textwidth}
     %\includegraphics[width=\linewidth]{rank.pdf}
            %\caption{\textbf{Numerical ranks} of all embeddings $\Z_\mathrm{\Theta}$  and embeddings from each ground-truth cluster $\cup_{\ell=1}^{10} \Z_\mathrm{\Theta}^{(\ell)}$ on CIFAR-10.}
    %\label{fig:rank}
     %\end{minipage}
     \caption{\text{Learning curves} of each loss term, ACC, and NMI during training. }
     \label{fig:obj-curves-ACC-NMI}
\end{figure}

\myparagraph{Learning curves}
To show the effectiveness of the two-stage training procedure, we display 
%\subsection{Learning curves}
%We visualize 
the learning curves of the proposed CgMCR$^2$ on CIFAR-20, where
%
% Learning curves on other datasets can be found in the supplementary.
%
we use 10 epochs (each epoch contains 33 mini-batch iterations) for warm-up and another 40 epochs for fine-tunning.
% 
%\myparagraph{Objective}
%To validate the effectiveness of our training strategy, 
%In Figs.~\ref{fig:obj_curve} and~\ref{fig:acc_curve}, 
In Fig.~\ref{fig:obj-curves-ACC-NMI}, 
we show the learning curve of each term in the objective of CgMCR$^2$, \ie, $R(\Z_\mathrm{\Theta};\epsilon)$, $R_c (\Z_\mathrm{\Theta}, \mathbf{\Pi}_\mathrm{\Phi};\epsilon)$ and $\L_\mathrm{Ncut}(\mathbf{\Pi}_\mathrm{\Phi};\mathbf{A},\gamma)$, ACC and NMI during the training. 
%defined in %by Eq.~
%\eqref{eq:cgmcr2}.
%
As can be observed in %Figs.~\ref{fig:obj_curve} and and~\ref{fig:acc_curve} 
Fig.~\ref{fig:obj-curves-ACC-NMI} that, the initialization stage did perform a %the pretext task in 
good warm-up, and during the fine-tuning stage, the value of $R(\Z_\mathrm{\Theta};\epsilon)$ remains nearly constant, but $R_c (\Z_\mathrm{\Theta}, \mathbf{\Pi}_\mathrm{\Phi};\epsilon)$ is rapidly optimized toward to its minimum. 
Moreover, the curves of ACC and NMI improve rapidly and achieves the optimal clustering results with the self-supervision of well-initialized cluster membership in the fine-tuning stage.

\myparagraph{Evaluation on Different Training Strategy} 
%
%We provide the ablation study on CIFAR-10, CIFAR-20 and CIFAR-100 datasets to evaluate the effect of the proposed objective and training strategy.
%
To validate the effectiveness of our two-stage training procedure, we conduct experiments on CIFAR-10 and CIFAR-100.  
Experimental results are shown in Table~\ref{tab:ablation_strategy}. 
%we optimize the objectives of MCR$^2$ and CgMCR$^2$ directly or by employing our two-stage training strategy.
%All other parameters are fixed to their optimal values.
%
The one-shot initialization learned by the pretext task (\ie, $-R+\L_\mathrm{Ncut}$) evidently improves the performance of optimizing both MCR$^2$ (\ie, $-R+R_c$) and CgMCR$^2$ (\ie, $-R+R_c+\L_\mathrm{Ncut}$) objectives in subsequent fine-tuning.
We can see that combining the pretext-task with MCR$^2$ objective also serves as a competitive baseline, demonstrating the importance of a principled initialization for $\mathbf{\Pi}_\mathrm{\Phi}$.
The proposed CgMCR$^2$ objective also demonstrates better empirical performance compared to the MCR$^2$ objective, regardless of the training strategy employed.

\begin{table}[t]
\centering
\caption{Evaluation on Different Training Strategy 
%\textbf{Ablation study of training strategy %and objective
on CIFAR-10 and CIFAR-100. We report ACC and NMI of the cluster head outputs.
}
\setlength{\tabcolsep}{2mm}{
    \begin{tabular}{c|c|cccc}
    \toprule
    \multirow{2}{*}{Warm-up} & \multirow{2}{*}{Fine-tune} & \multicolumn{2}{c}{CIFAR-10} & \multicolumn{2}{c}{CIFAR-100}  \\
    &  & ACC & NMI & ACC & NMI \\
    \midrule
     N/A & $-R+R_c$ & 86.5 & 86.8 & 60.4 & 67.4 \\
     N/A & $-R+R_c + \L_\mathrm{Ncut}$ & 91.2 & 89.3 & 66.0 & 70.8 \\
     $-R+\L_\mathrm{Ncut}$ & $-R+R_c$ & \underline{97.3} & \underline{93.9} & \underline{72.8} & \underline{79.8} \\
     $-R+\L_\mathrm{Ncut}$ & $-R+R_c + \L_\mathrm{Ncut}$ & \bf 97.7 & \bf 94.3 & \bf 77.8 & \bf 82.2 \\
    \bottomrule
    \end{tabular}
}
\label{tab:ablation_strategy}
\end{table}

\myparagraph{Evaluation on Sensitivity of Hyper-parameters} 
To evaluate the sensitivity of the performance of our CgMCR$^2$ to the hyper-parameters, we conduct experiments on CIFAR-10 and CIFAR-100. 
We % fix other parameters and 
vary the value of hyper-parameters $\epsilon$ and $\gamma$ in~\eqref{eq:cgmcr2} and report the clustering results. %on CIFAR-10 and CIFAR-100.
We set $\epsilon$ in the range of $\{0.2,0.3,0.4,0.5\}$ while $\gamma$ in a wide range of feasible region.\footnote{We follow the practical method in~\cite{He:PR2024-neuncut} to find the feasible $\gamma$ without using the ground-truth labels.} %annotations.}. 
In Fig.~\ref{fig:ablation_hyper} we show the experimental results. As can be observed that our method is less sensitive to the hyper-parameters $\epsilon$ and $\gamma$ % is ? and ?.
Then, we fix the hyper-parameters $\epsilon$ and $\gamma$ %i to ? an ?, 
and conduct experiments with varying 
%We proceed by varying the 
model parameters of the feature head $f(\cdot;\mathrm{\Theta})$ and the cluster head $g(\cdot;\mathrm{\Phi})$, where 
we set the number of hidden layers as $\{1,2,3,4\}$ and the number of neurons in each hidden layer as $\{512,1024,2048,4096\}$. 
Experimental results are shown in 
Fig.~\ref{fig:ablation_model}. Again, we can see that % shows that 
CgMCR$^2$ is not sensitive to the model size whenever it contains at least 1 hidden layers with 1024 hidden neurons.

\begin{table}
\centering
\caption{\textbf{Evaluation on Other Choices for %Effect of the 
Clustering Module on CIFAR-10, -20, -100 and TinyImageNet.} `s/it' denotes seconds per training iteration.}
    \begin{tabular}{c|c|cccccccc}
    \toprule
    \multirow{2}{*}{Clustering module} & \multirow{2}{*}{Train time (s/it)} & \multicolumn{2}{c}{CIFAR-10} & \multicolumn{2}{c}{CIFAR-20} & \multicolumn{2}{c}{CIFAR-100} & \multicolumn{2}{c}{TinyImageNet} \\
    & & ACC & NMI & ACC & NMI & ACC & NMI & ACC & NMI \\
    \midrule
     $g(\cdot;\mathrm{\Phi})$ & 0.02 & \textbf{97.7} & \textbf{94.3} & \textbf{68.8} & \textbf{74.0} & \textbf{78.3} & \textbf{82.5} & \textbf{72.9} & \textbf{81.4}  \\
     $k$-means & 6.8 & 96.8 & 92.5 & 53.6 & 66.4 &  74.2 & 80.1 & 68.5 & 79.7\\
     Spectral & 4.9 & 97.1 & 93.0 & 60.2 & 70.9 & \textbf{78.3} & \underline{82.4} & \underline{71.9} & \underline{80.8} \\
     {\mcb EnSC } & 11.2 & \underline{97.3} & \underline{93.4} &\underline{64.4} & \underline{71.3} & 76.7 & 81.7 & 68.7 & 80.6 \\
     
    \bottomrule
    \end{tabular}
\label{tab:cluster_module}
\end{table}

\begin{figure}[ht]
\centering
\begin{subfigure}{0.45\linewidth}
    \includegraphics[width=\textwidth,height=0.7\textwidth]{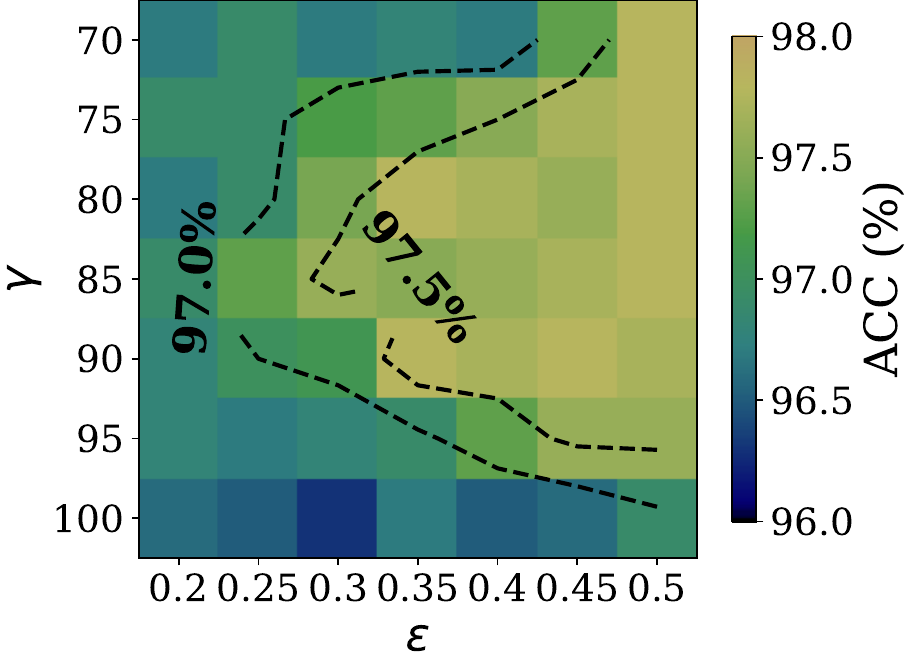}
\end{subfigure} 
\begin{subfigure}{0.45\linewidth}
    \includegraphics[width=\textwidth,height=0.7\textwidth]{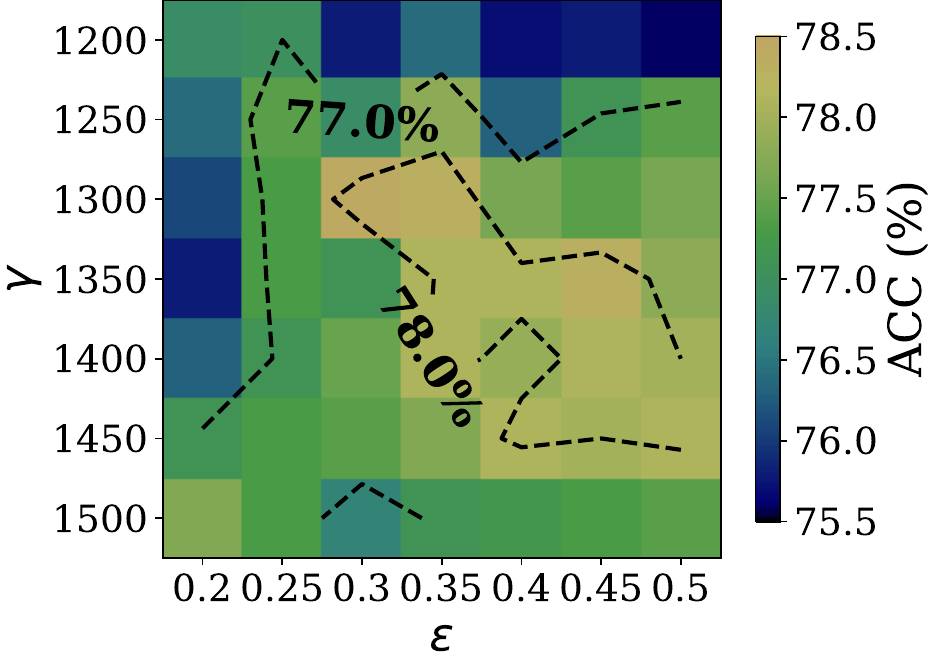}
\end{subfigure}
\caption{{\mcb \textbf{Effect of hyper-parameters} on CIFAR-10 (\textit{left}) and CIFAR-100 (\textit{right}).}}
\label{fig:ablation_hyper}
\end{figure}

\begin{figure}[ht]
\centering
\begin{subfigure}{0.45\linewidth}
    \includegraphics[width=\textwidth,height=0.7\textwidth]{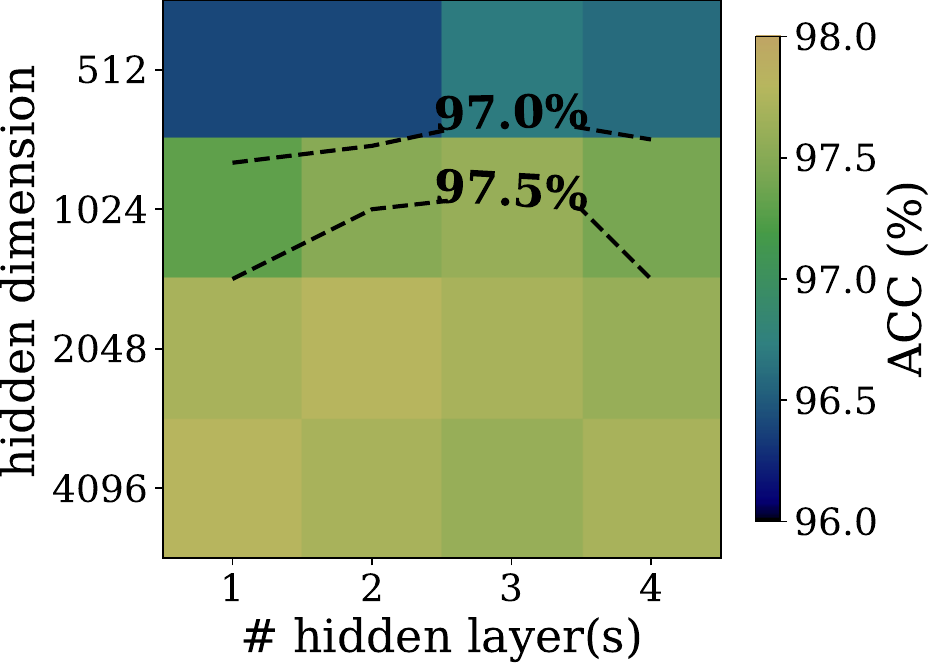}
\end{subfigure}
\begin{subfigure}{0.45\linewidth}
    \includegraphics[width=\textwidth,height=0.7\textwidth]{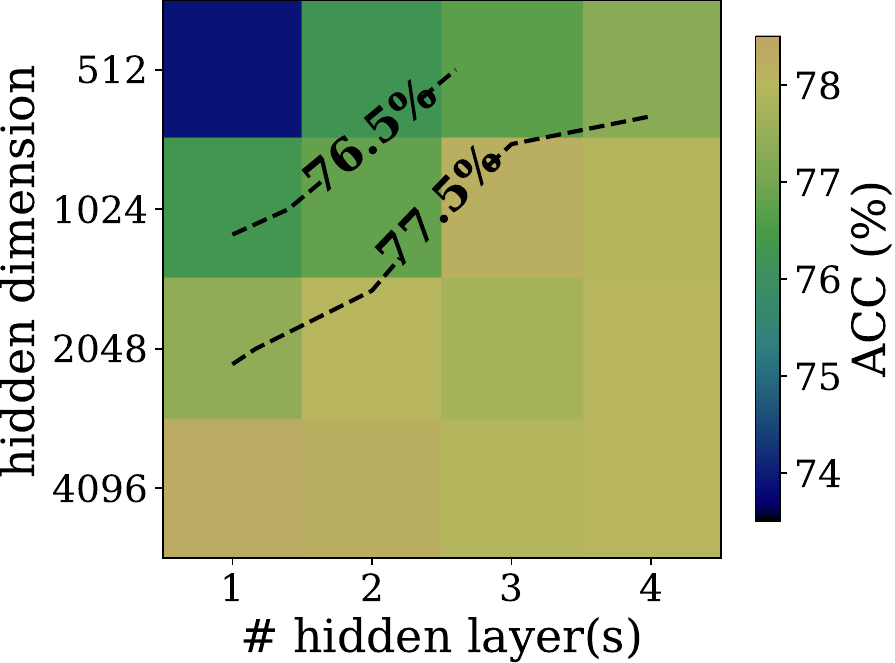}
\end{subfigure}
\caption{{\mcb \textbf{Effect of model parameters} on CIFAR-10 (\textit{left}) and CIFAR-100 (\textit{right}).}}
\label{fig:ablation_model}
\end{figure}

%\myparagraph{Ablation on the clustering module}
\myparagraph{Evaluation on Other Choices for Clustering Module}
Recall that the core idea of our CgMCR$^2$ is to %learn 
introduce a clustering module to produce the partition that is able to dynamically guide the feature learning via MCR$^2$.
%
%In Table~\ref{tab:cluster_module}, 
Here, we replace the differential spectral clustering module (\ie, $g(\cdot;\mathrm{\Phi})$) of CgMCR$^2$ with the conventional $k$-means~\cite{MacQueen-1967}, Spectral Clustering via Ncut~\cite{Shi:IEEE2000-Ncut} and {\mcb Elastic Net Subspace Clustering (EnSC)~\cite{You:CVPR2016-EnSC}} individually, to form corresponding learning frameworks. %and report the corresponding clustering performance.
Specifically, we use the conventional clustering methods to produce the cluster membership $\mathbf{\Pi}$ in each mini-batch training iteration to guide the fine-tunning of $\Z_\mathrm{\Theta}$.
We report the time cost and the clustering accuracy %Experimental results are reported 
in Table~\ref{tab:cluster_module}. 
%
%As can be seen, 
We can read that, using the cluster membership produced by conventional clustering algorithms to %for guiding 
guide the feature learning still yield very competitive clustering accuracy, in most cases. Nevertheless, 
%outperforms all other CLIP-based and MCR$^2$-based clustering methods except for CgMCR$^2$ on CIFAR-100 and TinyImageNet. 
% which demonstrates that our idea is reasonable and empirical effective, yet it has not been proposed in existing clustering methods based on MCR$^2$.
%
%However, the training 
the time cost of using conventional clustering algorithms is 100$\times$ %slower 
expensive than that of using the cluster head $g(\cdot;\mathrm{\Phi})$ due to the inability to utilize GPU acceleration, and can not infer the unseen data points in test sets directly.
%
%Therefore, we suggest 
This confirms the advantage of integrating a differential spectral clustering module $g(\cdot;\mathrm{\Phi})$ in the pipeline of CgMCR$^2$.

\section{Conclusion}
\label{sec:conclusion}
We have proposed a joint framework---graph Cut-guided Maximal Coding Rate Reduction---for learning the structured embeddings and clustering both in principled way. To be specific, a differential spectral clustering module is employed to learn the clustering membership to guide the task of learning structured embeddings.
We have conducted extensive experiments on five benchmark datasets and shown state-of-the-art clustering performance. In addition, we have also provided a set of ablation studies to validate the effectiveness of each component. 
%
%Still, we observe the simplicity in architecture of our framework and the potential to improve the cluster module, which are left for future work.

\begin{credits}
\subsubsection{\ackname} This study was funded by the National Natural Science Foundation of China under Grant No.~61876022. 

%Chun-Guang Li is the corresponding author.

\subsubsection{\discintname}
The authors declare that they have no known competing financial interests or personal relationships that could have appeared to influence the work reported in this paper.
\end{credits}

}

%
% ---- Bibliography ----
\bibliographystyle{splncs04}
\bibliography{weihe}

\newpage
\appendix
\renewcommand{\thefigure}{\Alph{section}.\arabic{figure}}
\renewcommand{\thetable}{\Alph{section}.\arabic{table}}
\setcounter{figure}{0}
\setcounter{table}{0}
\renewcommand{\appendixname}{Appendix~\Alph{section}}
\begin{sc}
\begin{center}
    \Large Supplementary Materials
\end{center}
\end{sc}

\section{Experimental Details}

\myparagraph{Dataset description}
In Table~\ref{tab:data}, we provide an overview of of all selected datasets.
The images in MNIST and F-MNIST are grayscale, and images of all datasets are resized to $224\times 224$ dimensions to serve as the inputs of CLIP image encoder.
For Oxford Flowers-102 and Stanford Dogs-120, we train and test our CgMCR$^2$ on the entire dataset.
For all other datasets, we use the train set and test set for training and testing, respectively.

\begin{table}[h]
\small
\caption{\textbf{Specification of all selected datasets}.}
    \centering
    \begin{tabular}{lccc}
        \toprule
        Dataset & \# Classes & \# Training & \# Testing \\
        \midrule
        MNIST         & 10  & 60,000 & 10,000 \\
        F-MNIST       & 10  & 60,000 & 10,000 \\
        CIFAR-10      & 10  & 50,000 & 10,000 \\
        CIFAR-20      & 20  & 50,000 & 10,000 \\
        CIFAR-100     & 100 & 50,000 & 10,000 \\
        Flowers-102   & 102 & 8,192  & N/A \\
        Dogs-120      & 120 & 20,580 & N/A \\
        TinyImageNet  & 200 & 100,000 & 10,000 \\
        ImageNet-1k & 1000 & 1,281,167 &  50,000 \\
        \bottomrule
    \end{tabular}
    \label{tab:data}
\end{table}

\begin{table}[h]
    \caption{\textbf{Model parameters} of the pre-feature layer, feature head, and cluster head (from left to right).}
    \begin{subtable}[t]{0.32\linewidth}
    \centering
        \begin{tabular}{c}
            \toprule
            Linear: $\RR^{768}\rightarrow \RR^{4096}$ \\
            \midrule
            BatchNorm1d($4096$) \\
            \midrule
            ReLU \\
            \bottomrule
        \end{tabular}
    \end{subtable}
    \hfill
    \begin{subtable}[t]{0.32\linewidth}
    \centering
        \begin{tabular}{c}
            \toprule
            Linear: $\RR^{4096}\rightarrow \RR^{4096}$ \\
            \midrule
            ReLU \\
            \midrule
            Linear: $\RR^{4096}\rightarrow \RR^{d}$ \\
            \bottomrule
        \end{tabular}
    \end{subtable}
    \hfill
    \begin{subtable}[t]{0.32\linewidth}
    \centering
        \begin{tabular}{c}
            \toprule
            Linear: $\RR^{4096}\rightarrow \RR^{4096}$ \\
            \midrule
            ReLU \\
            \midrule
            Linear: $\RR^{4096}\rightarrow \RR^{k}$ \\
            \midrule
            Gumbel-Softmax \\
            \bottomrule
        \end{tabular}
    \end{subtable}
    \label{tab:arch}
\end{table}

\myparagraph{Parameters for CgMCR$^2$}
In Table \ref{tab:arch}, we detail the model parameters of our framework.
In Table~\ref{tab:param}, we detail the optimal hyper-parameters for CgMCR$^2$.
% Generally speaking, a larger number of ground-truth clusters necessitates a larger batch size.
The proposed CgMCR$^2$ demonstrates robustness to variations in batch size, $\gamma$ and $\epsilon$.
Typically, employing a larger batch size along with a higher learning rate tends to yield more stable performance.
Meanwhile, CgMCR$^2$ with larger batch size requires more training iteration to converge.
\begin{table}[t]
\centering
\caption{\textbf{Optimal hyper-parameters.}``lr'' and ``wd'' are the learning rate and weight decay of Adam optimizer, $d$ is the output dimension of feature head, $T_1$ denotes warm-up epochs, $T_2$ denotes fine-tuning epochs, $\gamma$ and $\epsilon$ are the hyper-parameters of CgMCR$^2$ objective, and $s$ is the number of nonzero affinity entries kept in each row.}
\setlength{\tabcolsep}{1mm}{
    \begin{tabular}{l |ccccccccc}
    \toprule
    Dataset & lr & wd & $d$ & $T_1$ & $T_2$ & bs & $\gamma$ & $\epsilon$ & $s$ \\
    \midrule
    MNIST        & 0.001  & 0.001  & 128 & 20 & 30 & 2048&  50 & 0.5 & 20 \\
    F-MNIST      & 0.001  & 0.001  & 128 & 20 & 30 & 2048&  50 & 0.2 & 20 \\
    CIFAR-10     & 0.0001 & 0.0005 & 128 & 10 & 10 & 512 &  70 & 0.5 & 10 \\
    CIFAR-20     & 0.0001 & 0.0005 & 128 & 10 & 40 &1500 &  80 & 0.2 & 50 \\
    CIFAR-100    & 0.0005 & 0.0001 & 128 & 20 & 30 &2048 & 1400& 0.5 & 20 \\
    Flowers-102  & 0.0005 & 0.0005 & 128 & 20 & 30 &2048 & 1200 & 0.5 & 10 \\
    Dogs-120     & 0.001  & 0.001  & 128 & 20 & 30 &2048 & 1100 & 0.2 & 40 \\ 
    TinyImageNet & 0.0003 & 0.0005 & 256 & 20 & 30 & 2048 & 3000 & 0.5 & 20 \\
    ImageNet & 0.001  & 0.0001 & 256 & 10 & 10 & 3000 & 50000 & 0.2 & 3 \\ 
    \bottomrule
    \end{tabular}
}
\label{tab:param}
\end{table}

\myparagraph{Searching parameters for clustering methods}
When comparing with classical clustering methods and reproduced deep clustering methods, we report their best performance through a greedy search for optimal parameters, as shown in Table~\ref{tab:other_param}.
In Spectral Clustering, $\sigma$ serves as the bandwidth parameter of the Gaussian kernel, and we reserve the $s$ largest entries of each row in the affinity matrix.
In EnSC, $\tau\in[0,1]$ is the parameter regulating the sparsity of self-expressive coefficients and $\beta$ is the trade-off parameter balancing the self-expressive error against the sparsity regularizer.
In SCAN, $\mu$ is the weight of the between-cluster entropy-maximizing regularization.
\begin{table}[t]
    \centering
    \caption{\textbf{Parameter search} with the following parameters for Spectral Clustering, EnSC and SCAN.}
    \begin{tabular}{l|c}
        \toprule
         Method & Search scope for parameters \\
         \midrule
         Spectral Clustering & $\sigma\in \{3,2,1,0.5,0.4,0.3,0.2,0.1,0.07,0.05\}$, $s\in\{3,10,100,1000\}$\\
         EnSC & $\tau \in \{0.9,0.95,1\}$, $\beta\in\{1,2,5,10,50,100,200\}$ \\
         SCAN & $\mu\in\{1,2,4,10,20,50,100,200,500,1000,2000\}$ \\
         \bottomrule
    \end{tabular}
    \label{tab:other_param}
\end{table}

\myparagraph{The MoCo pre-trained model}
To train our CgMCR$^2$ from scratch, we leverage MoCo-v2, a self-supervised learning method, to learn pre-features.
% For a fair comparison to other methods, we use ResNet18 as the backbone of pre-trained MoCo-v2.
%
The MoCo-v2 image encoder takes two augmentations of each image as inputs, and we utilize the averaged output embedding of the two augmentations as the pre-feature.
The augmentation strategy follows that in NMCE, and is detailed in Table~\ref{tab:aug}.

\begin{table}[t]
    \centering
    \caption{\textbf{Augmentation strategy} of MoCo-v2.}
    \begin{tabular}{l}
        \toprule
        \texttt{from torchvision.transforms import *} \\
        \midrule
        \texttt{Compose([} \\
        \texttt{\quad RandomResizedCrop(32,scale=(0.08, 1.0)), } \\
        \texttt{\quad RandomHorizontalFlip(p=0.5), } \\
        \texttt{\quad RandomApply([ColorJitter(0.4, 0.4, 0.4, 0.1)], p=0.8), } \\
        \texttt{\quad RandomGrayscale(p=0.2), } \\
        \texttt{\quad ToTensor(),} \\
        \texttt{\quad Normalize([0.4914, 0.4822, 0.4465], [0.2023, 0.1994, 0.2010])} \\
        \quad \texttt{)])} \\
        \bottomrule
    \end{tabular}
    \label{tab:aug}
\end{table}

\myparagraph{The CLIP pre-trained model}
CLIP is a large-scale language-supervised learning method that learns general semantic meaning from over 400 million text-image pairs.
In our approach, we utilize only the image encoder of the pre-trained CLIP model.
Images are resized to $224$ along the smaller edge and center-cropped to $224\times224$ before being inputted to the CLIP image encoder.
Subsequently, the features extracted by the CLIP image encoder are used as pre-features for our CgMCR$^2$.

\section{More Experiment Results}
\subsection{Ablation Study}

\myparagraph{Ablation on the output activation}
In our method, we employ the Gumbel-Softmax as the output activation function of the cluster head.
In Table~\ref{tab:ablation_softmax}, we compare the use of Softmax as the output activation function with the use of Gumbel-Softmax and report their respective best performances on CIFAR-10, -20, -100 and TinyImageNet.
As can be seen, the use of Gumbel-Softmax leads to slightly higher clustering accuracy on four standard datasets.
\begin{table}[t]
\setlength{\tabcolsep}{1mm}{
    \centering
    \caption{\textbf{Effect of output activation function.}}
    \begin{tabular}{l|cccccccc}
        \toprule
         \multirow{2}{*}{Output activation} & \multicolumn{2}{c}{CIFAR-10} & \multicolumn{2}{c}{CIFAR-20} & \multicolumn{2}{c}{CIFAR-100} & \multicolumn{2}{c}{TinyImageNet}  \\
          & ACC & NMI & ACC & NMI & ACC & NMI & ACC & NMI \\
         \midrule
         Softmax & 97.3 & 92.6 & 66.8 & 69.3 & 75.4 & 80.8 & 71.4 & 81.0 \\
         Gumbel-Softmax (Ours) &  97.7 & 94.3 & 68.8 & 74.0 & 78.3 & 82.5 & 72.9 & 81.4 \\
         \bottomrule
\end{tabular}
\label{tab:ablation_softmax}
}
\end{table}

\myparagraph{Ablation on the affinity}
We examine the effect of $\mathbf{A}$ with various definitions.
As described earlier, the affinity matrix in the proposed CgMCR$^2$ is defined by $\mathbf{A}\coloneqq\P_s(\Z_\mathrm{\Theta}^\top \Z_\mathrm{\Theta})$.
Following traditional spectral clustering approaches, we additionally use Gaussian kernel (a.k.a. the Radial Basis Function kernel) to define the affinities, \ie,

\begin{equation}
\label{eq:kernel}
    a_{i,j}=\exp \left(-\frac{\|\z_i-\z_j\|^2_2}{2\sigma^2}\right),
\end{equation}
where $\sigma$ is the bandwidth parameter.
In Table~\ref{tab:ablation_aff}, we report the best clustering performance on CIFAR-10, -100, training time per iteration and memory cost of using different affinity matrices.
All the experiments are conducted on a single NVIDIA GeForce 3080Ti GPU, and the batch size is set to 512 when recording the training time and memory cost.
\begin{table}[t]
    \centering
    \caption{\textbf{Varying definitions of $\mathbf{A}$ on CIFAR-10 an CIFAR-100.}}
    \begin{tabular}{l|cc|cccc}
        \toprule
         \multirow{2}{*}{Definition of $\mathbf{A}$} & \multirow{2}{*}{Time (ms/it)} & \multirow{2}{*}{Memory (MB)} & \multicolumn{2}{c}{CIFAR-10} & \multicolumn{2}{c}{CIFAR-100} \\
          & & & ACC & NMI & ACC & NMI \\
         \midrule
         Gaussian kernel & 23.2 & 2,309 & 97.5 & 93.8 & 75.8 & 81.7 \\
         Cosine similarity (Ours) & 22.9 & 2,030 & 97.7 & 94.3 & 78.3 & 82.5\\
         \bottomrule
    \end{tabular}
    \label{tab:ablation_aff}
\end{table}
As can be seen, computing the Gaussian kernel requires a bit higher computational and memory cost, and achieves slightly inferior performance compared to computing cosine similarity.

We proceed by evaluating the effect of different post-processing operator.
We notice that the doubly stochastic projection enjoys solid theoretical guarantees and \textit{state-of-the-art} performance as a post-processing method in subspace clustering~\cite{Lim:DSSC-arxiv20}.
Specifically, the doubly stochastic projection projects the affinity matrix onto a doubly stochastic space 
\begin{equation}
    \A\coloneqq\left\{\tilde{\mathbf{A}}\in\mathbb{R}^{N\times N}\mid\tilde{\mathbf{A}}\mathbf{1}=\mathbf{1},\tilde{\mathbf{A}}^\top\mathbf{1}=\mathbf{1}\right\}
\end{equation}
under the distance of a scaled $\mathbf{A}$:
\begin{equation}
    {\arg\min}_{\tilde{\mathbf{A}}\in\A} \left\| \tilde{\mathbf{A}} - \mu\mathbf{A}\right\|_F^2.
\end{equation}
This post-processing method also has been adopted in MLC and CPP.
In Table~\ref{tab:ablation_postprocess}, we use cosine similarity to define the affinity matrix and compare our method with doubly stochastic projection and the baseline with no post-processing.
In our framework, simply reserving $s$ largest entries of each row in $\mathbf{A}$ achieves the highest accuracy with almost no computational and memory cost,
while applying doubly stochastic projection produces less satisfactory clustering results and demands much more training time and GPU memory.
\begin{table}[t]
    \centering
    \caption{\textbf{Varying post-processing operators of $\mathbf{A}$ on CIFAR-10.}}
    \begin{tabular}{l|cc|cccc}
        \toprule
         \multirow{2}{*}{Post-processing of $\mathbf{A}$} & \multirow{2}{*}{Time (ms/it)} & \multirow{2}{*}{Memory (MB)} & \multicolumn{2}{c}{CIFAR-10} & \multicolumn{2}{c}{CIFAR-100} \\
          & & & ACC & NMI & ACC & NMI \\
         \midrule
         N/A & 22.4 & 1,905 & 96.7 & 91.6 & 70.1 & 78.6 \\
         Doubly stochastic & 74.1 & 5,465 & 97.2 & 92.4  & 75.1 & 80.8 \\
         Reserving top-$s$ entries (ours) & 22.9 & 2,030 & 97.7 & 94.3 & 78.3 & 81.9\\
         \bottomrule
        
    \end{tabular}
    \label{tab:ablation_postprocess}
\end{table}

\myparagraph{Ablation on parameter $s$}
We previously conducted an ablation study to evaluated the effect of hyper-parameters $\gamma$ and $\epsilon$.
Another important hyper-parameter is $s$, representing the number of entries reserved in each row of matrix $\mathbf{A}$. 
In this study, we proceed by evaluate the effect of varying $s$ on CIFAR-10 and CIFAR-100.
For CIFAR-10, we fix the batch size to 512 and report the clustering performance of CgMCR$^2$ with $s\in \{3,5,10,20,50,100,200,300,400,500\}$.
For CIFAR-100, we fix the batch size to 2048 and report the clustering performance of CgMCR$^2$ with $s\in \{3,5,10,20,50,100,200,300,400,500, 1000, 1500, 2000\}$.
As can be seen from Table~\ref{tab:ablation_s}, our method demonstrates robustness to the parameter $s$.
Specifically, values of $s$ within a wide range of $[5,500]$ yield satisfactory performance on CIFAR-10, while on CIFAR-100, values of $s$ within the range of $[10,1000]$ yield satisfactory performance.

\begin{table}[t]
    \centering
    \caption{\textbf{Clustering accuracy (\%)} of the CgMCR$^2$ with varying $s$ on CIFAR-10 and CIFAR-100.}
    \begin{tabular}{l|ccccccccccccc}
        \toprule
        \diagbox[height=1.7em]{Data}{$s$} & 3 & 5 & 10 & 20 & 50 & 100 & 200 & 300 & 400 & 500 & 1000 & 1500 & 2000 \\
        \midrule
        CIFAR-10 & 96.9 & 97.6 & 97.7 & 97.5 & 97.4 & 97.4 & 97.7 & 97.6 & 97.4 & 97.2 & - & - & - \\
        CIFAR-100 & 75.2 & 76.6 & 77.9 & 78.3 & 76.6 & 77.3 & 77.1 & 77.7 & 77.2 & 77.3 & 77.4 & 74.2 & 73.0\\
        \bottomrule
    \end{tabular}

    \label{tab:ablation_s}
\end{table}

\begin{figure}[t]
    \centering
    \begin{subfigure}[t]{0.3\linewidth}
        \includegraphics[width=\textwidth]{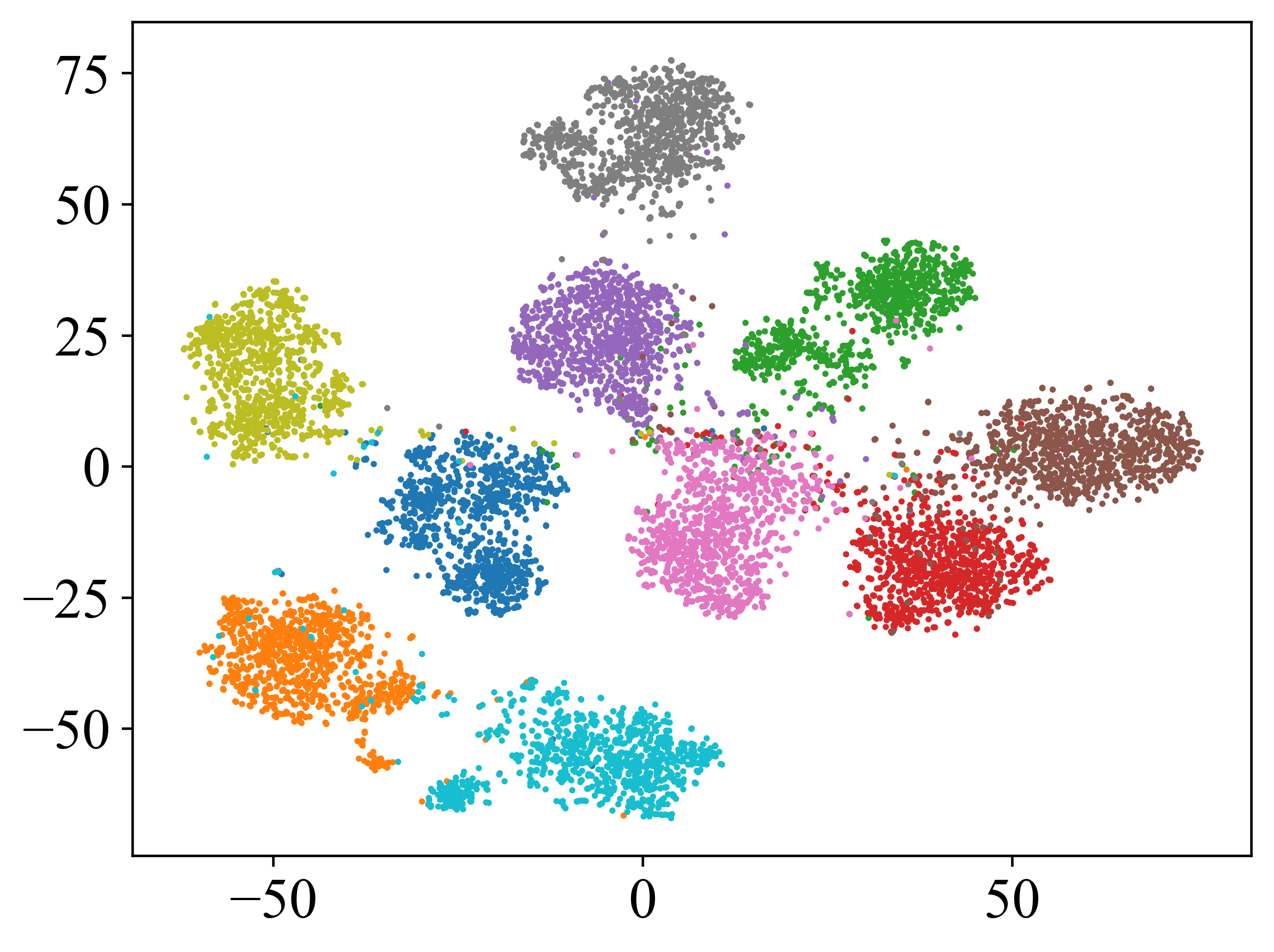}
        \caption{CLIP CIFAR-10}
    \end{subfigure}
    \begin{subfigure}[t]{0.3\linewidth}
        \includegraphics[width=\textwidth]{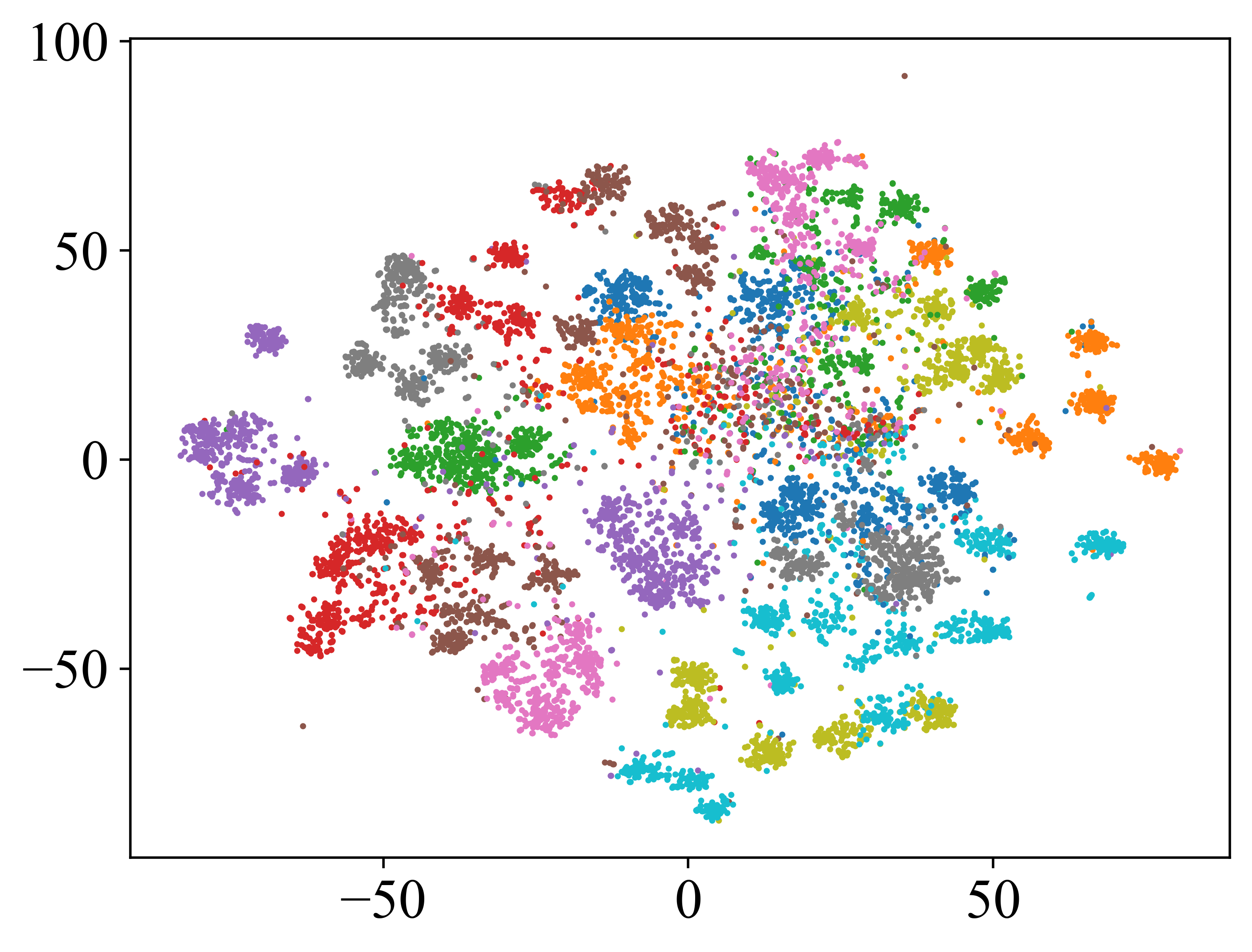}
        \caption{CLIP CIFAR-20}
        \label{subfig:clip_cifar20}
    \end{subfigure}
    \begin{subfigure}[t]{0.3\linewidth}
        \includegraphics[width=\textwidth]{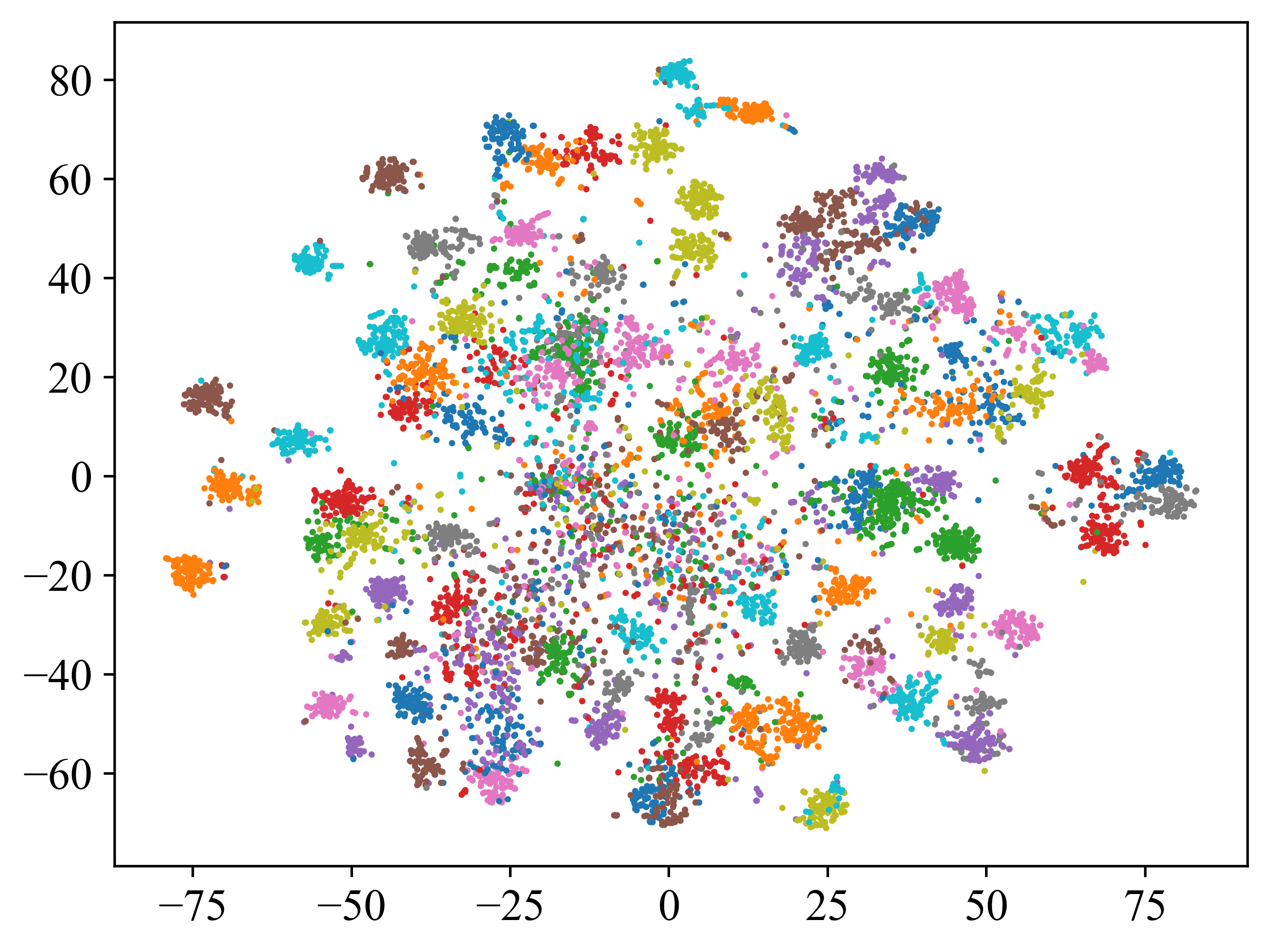}
        \caption{CLIP CIFAR-100}
    \end{subfigure} \\
    \begin{subfigure}[t]{0.3\linewidth}
        \includegraphics[width=\textwidth]{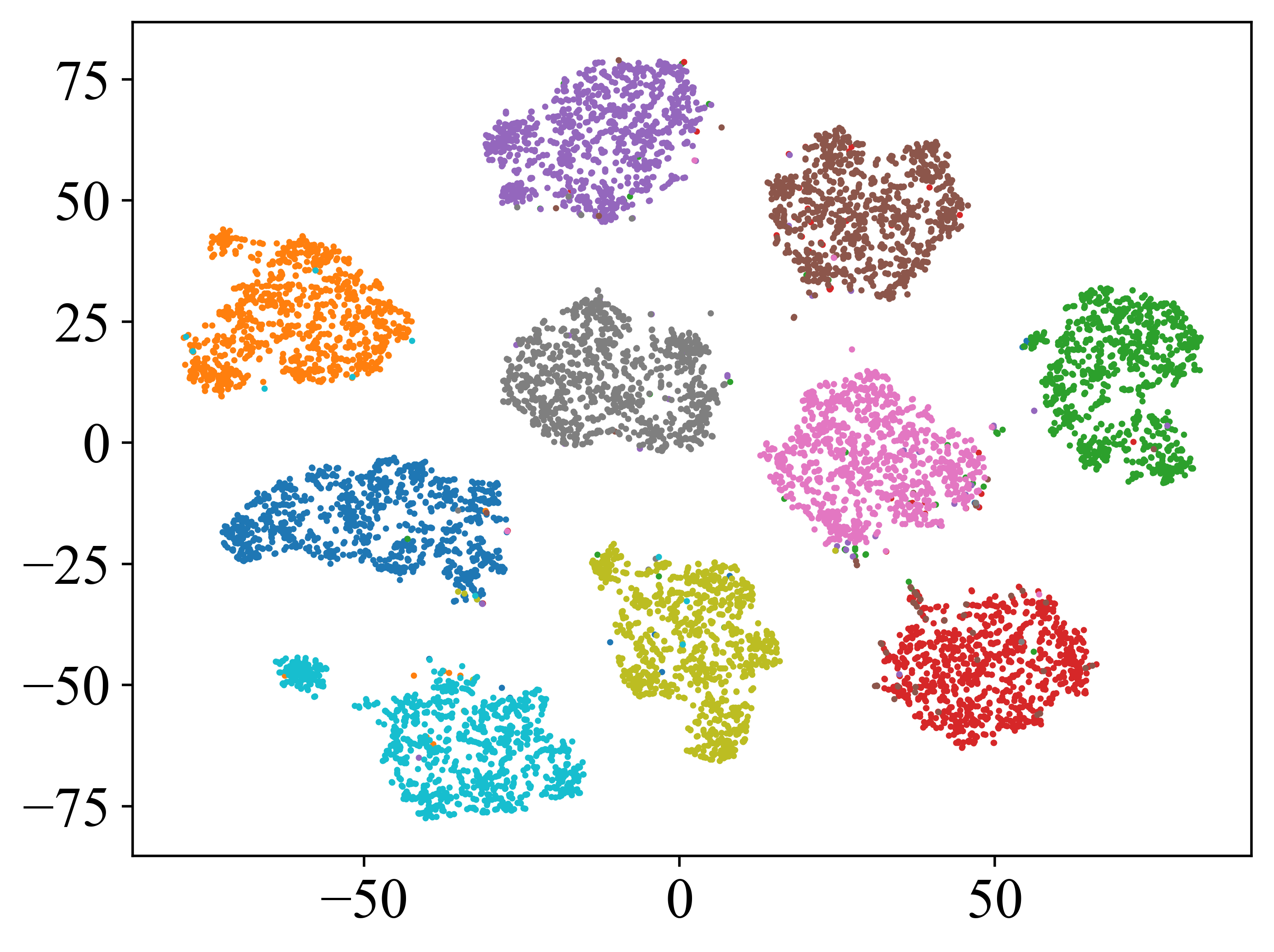}
        \caption{CgMCR$^2$ CIFAR-10}
    \end{subfigure}
    \begin{subfigure}[t]{0.3\linewidth}
        \includegraphics[width=\textwidth]{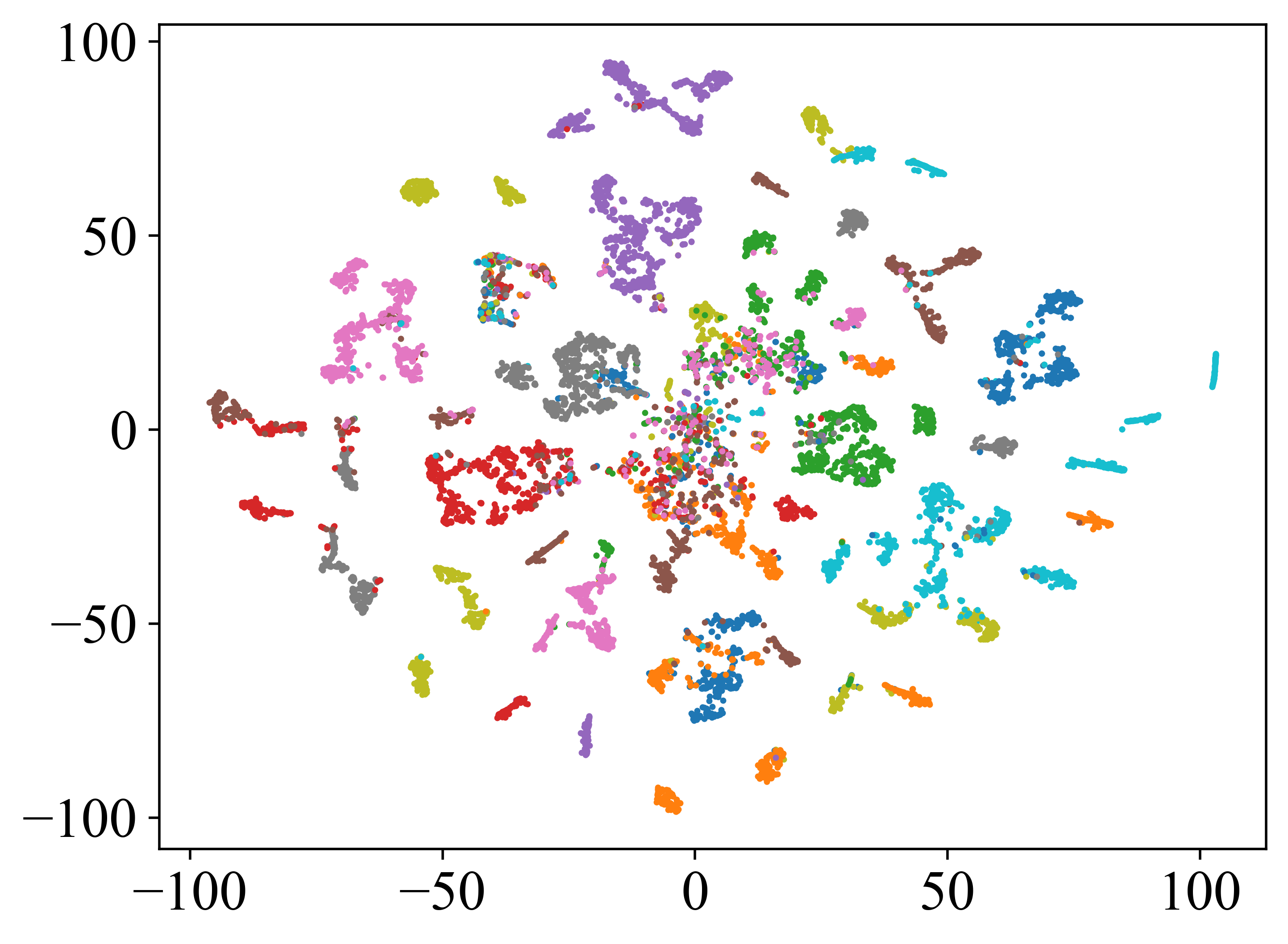}
        \caption{CgMCR$^2$ CIFAR-20}
        \label{subfig:sgmcr2_cifar20}
    \end{subfigure}
    \begin{subfigure}[t]{0.3\linewidth}
        \includegraphics[width=\textwidth]{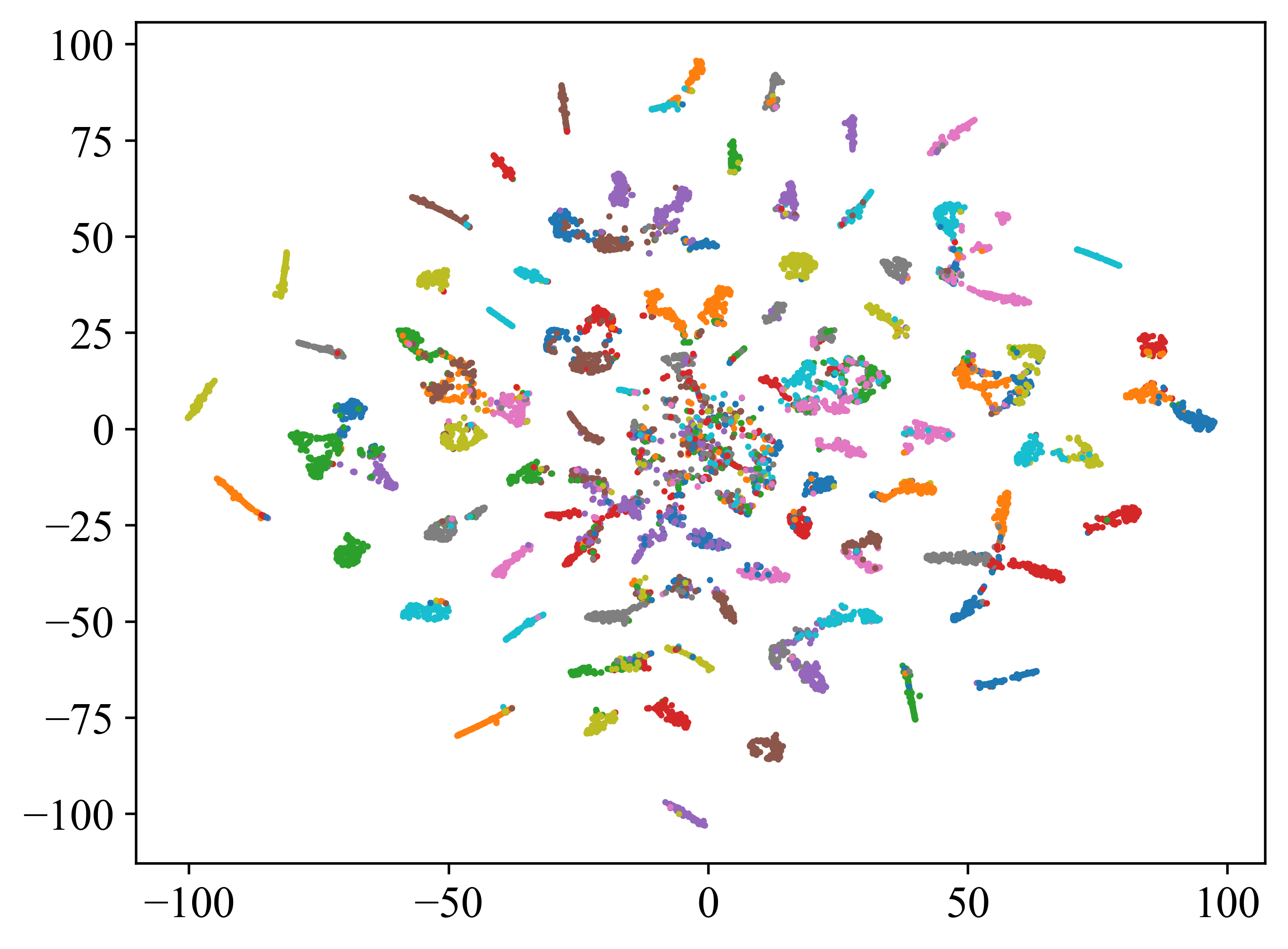}
        \caption{CgMCR$^2$ CIFAR-100}
    \end{subfigure}
\caption{\textbf{Utilizing t-SNE for 2-D visualization.} We plot \textbf{(a)--(c):} the CLIP pre-features on CIFAR-10, -20, -100 and \textbf{(d)--(f):} the CgMCR$^2$ features on CIFAR-10, -20, -100.}
\label{fig:tsne}
\end{figure}

\subsection{Visualization}

\myparagraph{Visualization via t-SNE}
To demonstrate the properties of representations learned by the feature head of CgMCR$^2$, we also utilize t-SNE to obtain 2-D visualization of the representations on CIFAR-10, CIFAR-20 and CIFAR-100.
In Fig.~\ref{fig:tsne}, it is evident that the proposed CgMCR$^2$ learns a more compact and discriminative representations from the CLIP features.

\begin{figure}[t]
\centering
    \begin{subfigure}[t]{0.32\linewidth}
        \includegraphics[width=\textwidth]{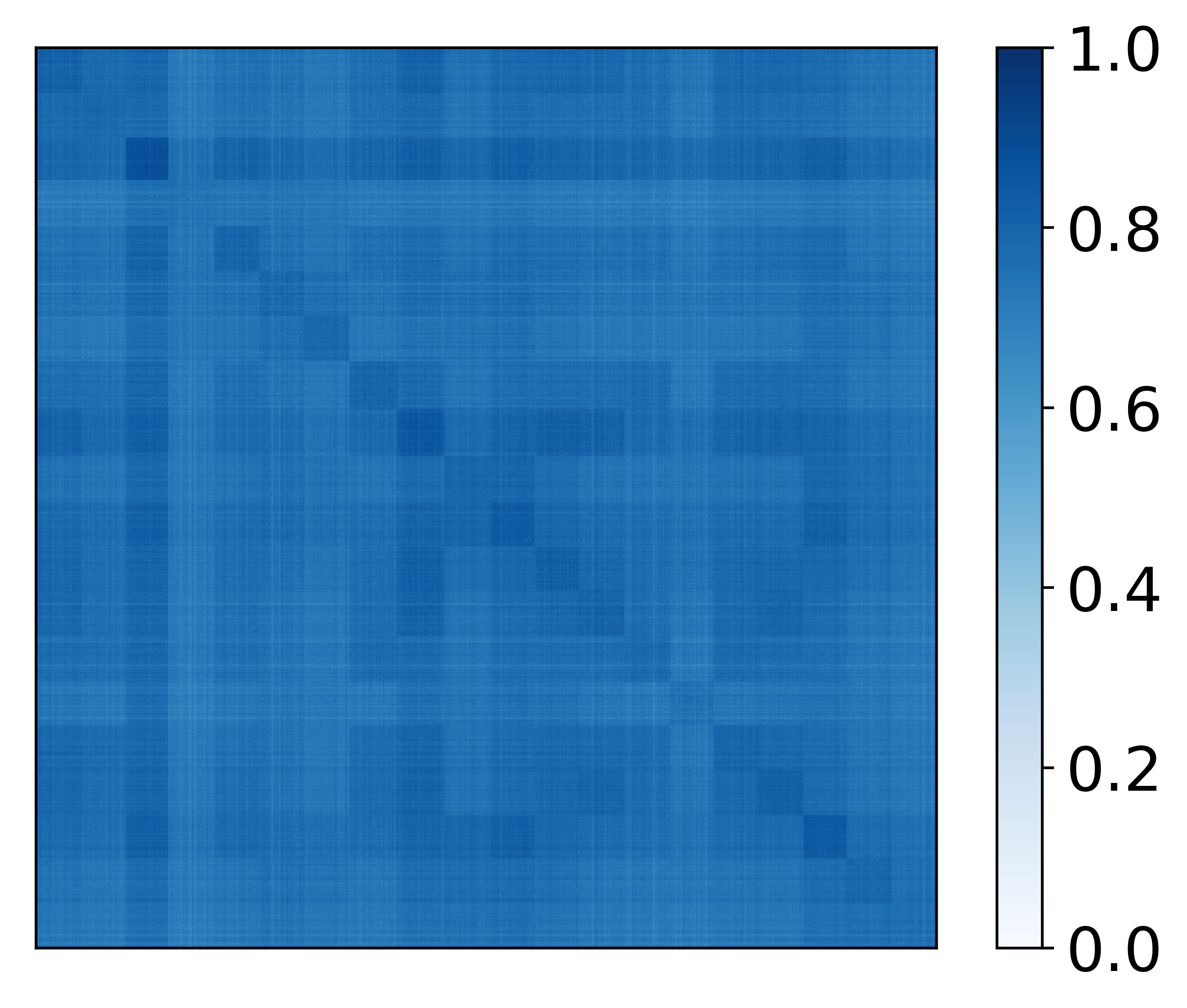}
        \caption{CLIP CIFAR-20}
    \end{subfigure}
    \begin{subfigure}[t]{0.32\linewidth}
        \includegraphics[width=\textwidth]{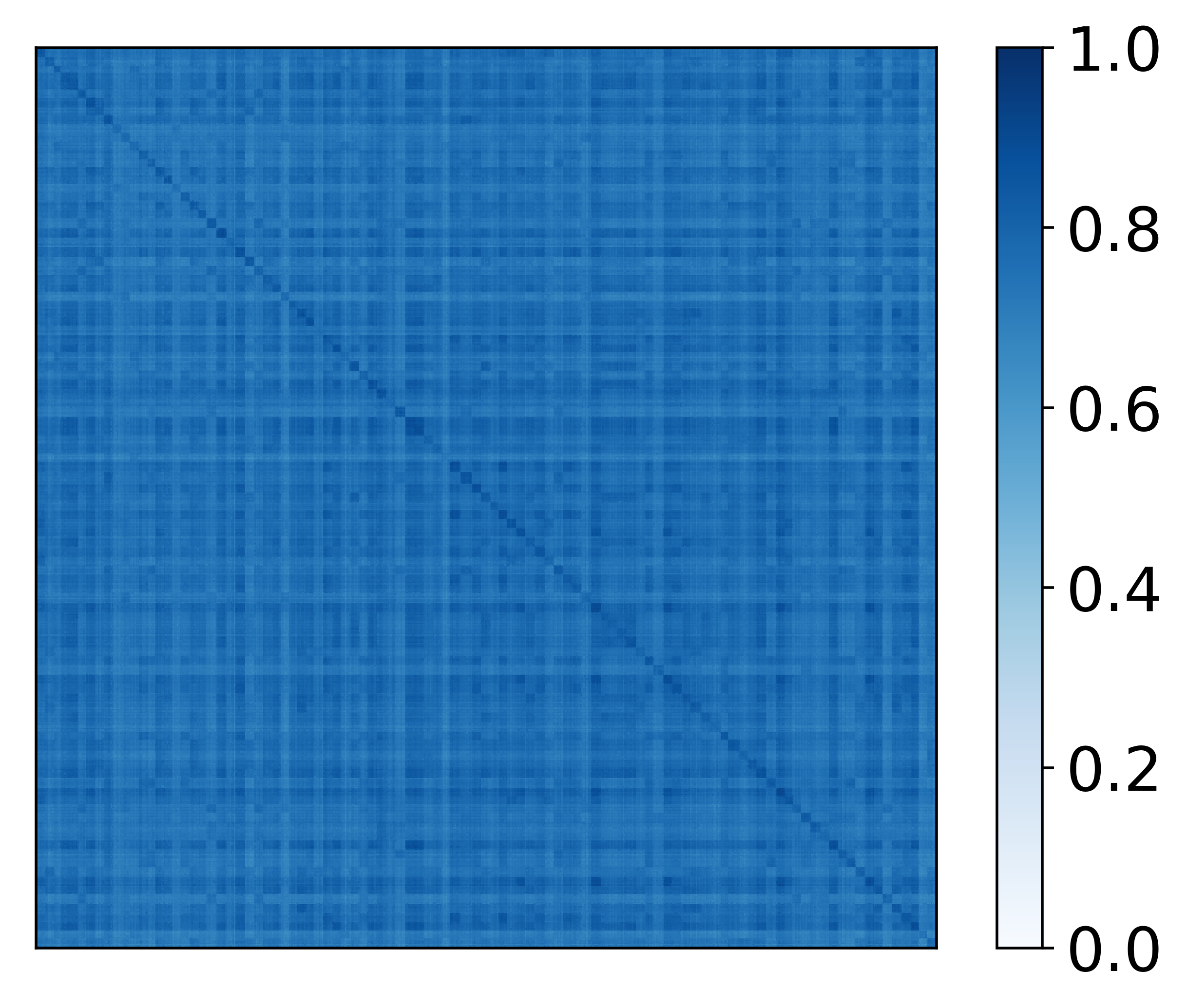}
        \caption{CLIP CIFAR-100}
    \end{subfigure}
    \begin{subfigure}[t]{0.32\linewidth}
        \includegraphics[width=\textwidth]{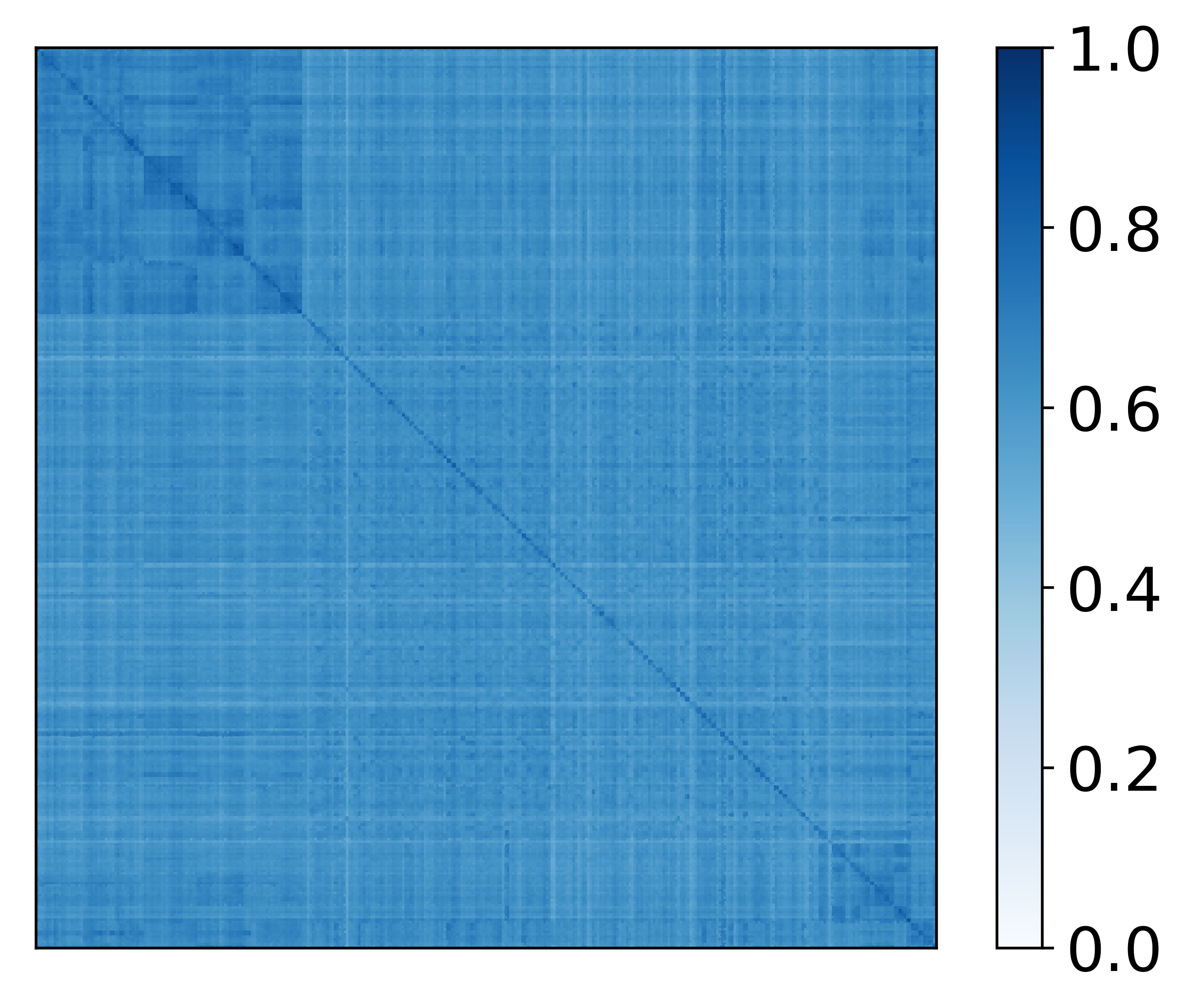}
        \caption{CLIP TinyImageNet}
    \end{subfigure} \\
    \begin{subfigure}[t]{0.32\linewidth}
        \includegraphics[width=\linewidth]{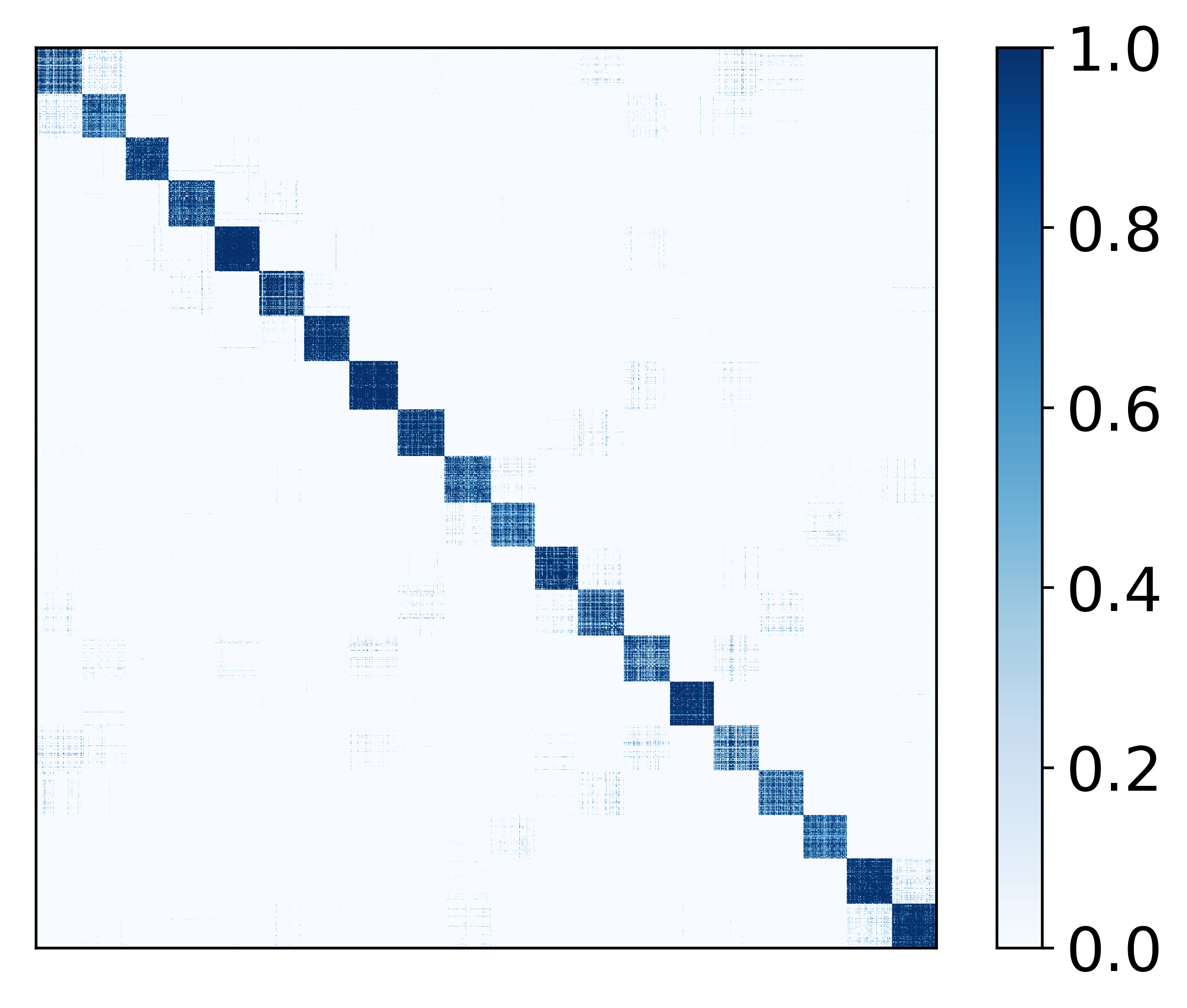}
        \caption{ $\mathbf{Z}_\mathrm{\Theta}$ CIFAR-20}
    \end{subfigure}
    \begin{subfigure}[t]{0.32\linewidth}
        \includegraphics[width=\linewidth]{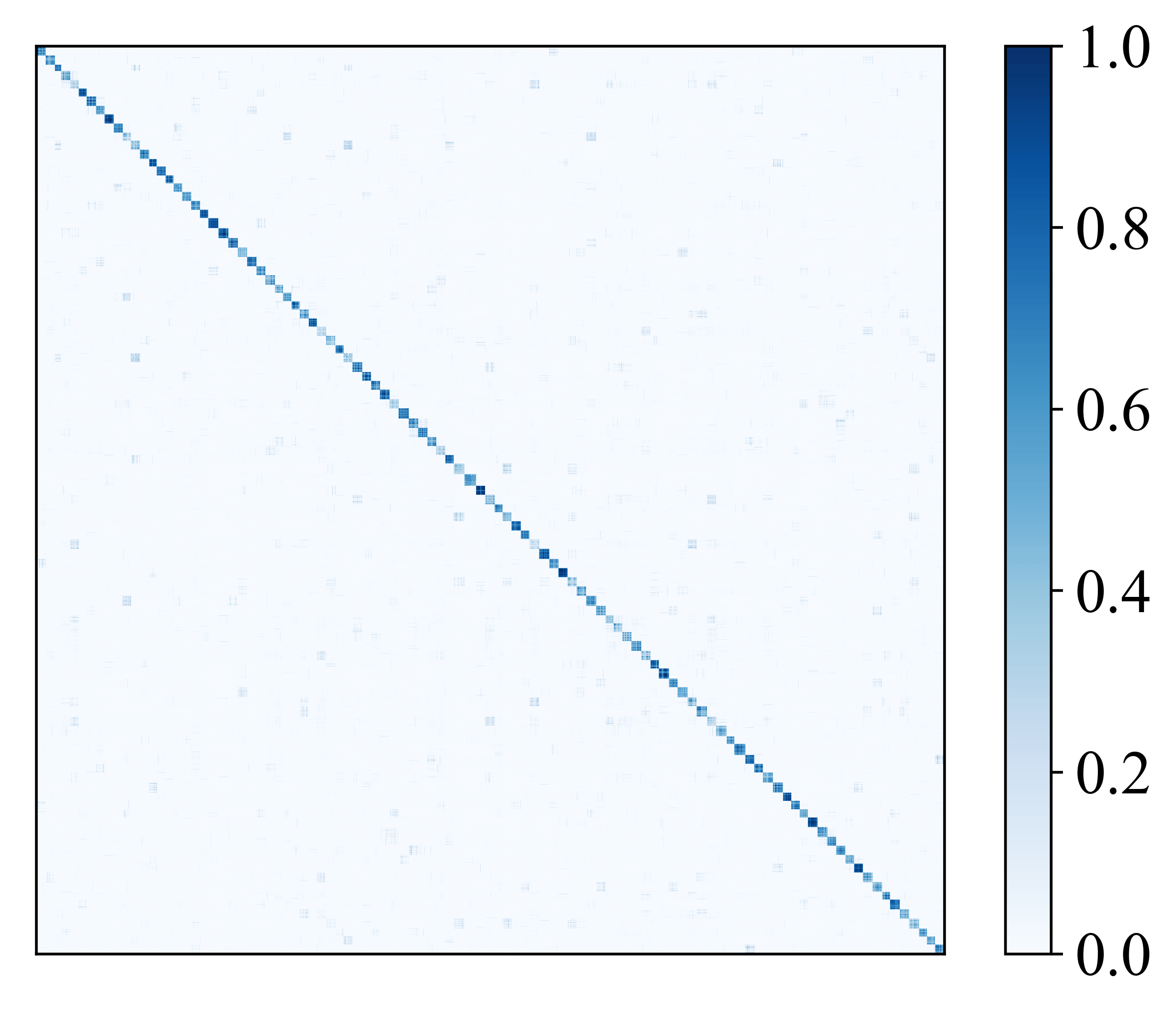}
        \caption{$\mathbf{Z}_\mathrm{\Theta}$ CIFAR-100}
    \end{subfigure}
    \begin{subfigure}[t]{0.32\linewidth}
        \includegraphics[width=\linewidth]{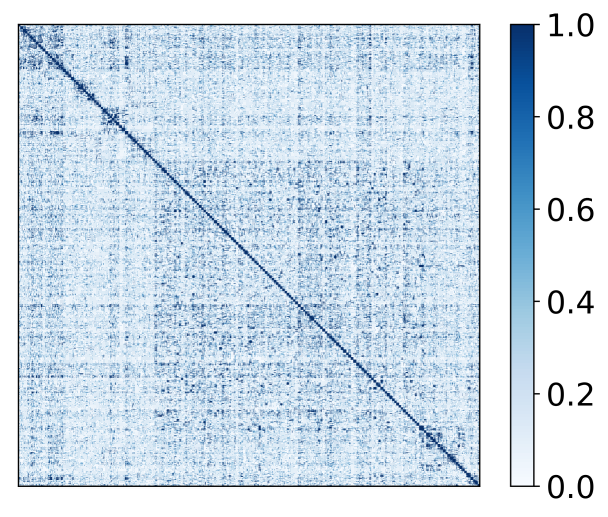}
        \caption{$\mathbf{Z}_\mathrm{\Theta}$ TinyImageNet}
    \end{subfigure} \\
    \begin{subfigure}[t]{0.32\linewidth}
        \includegraphics[width=\linewidth]{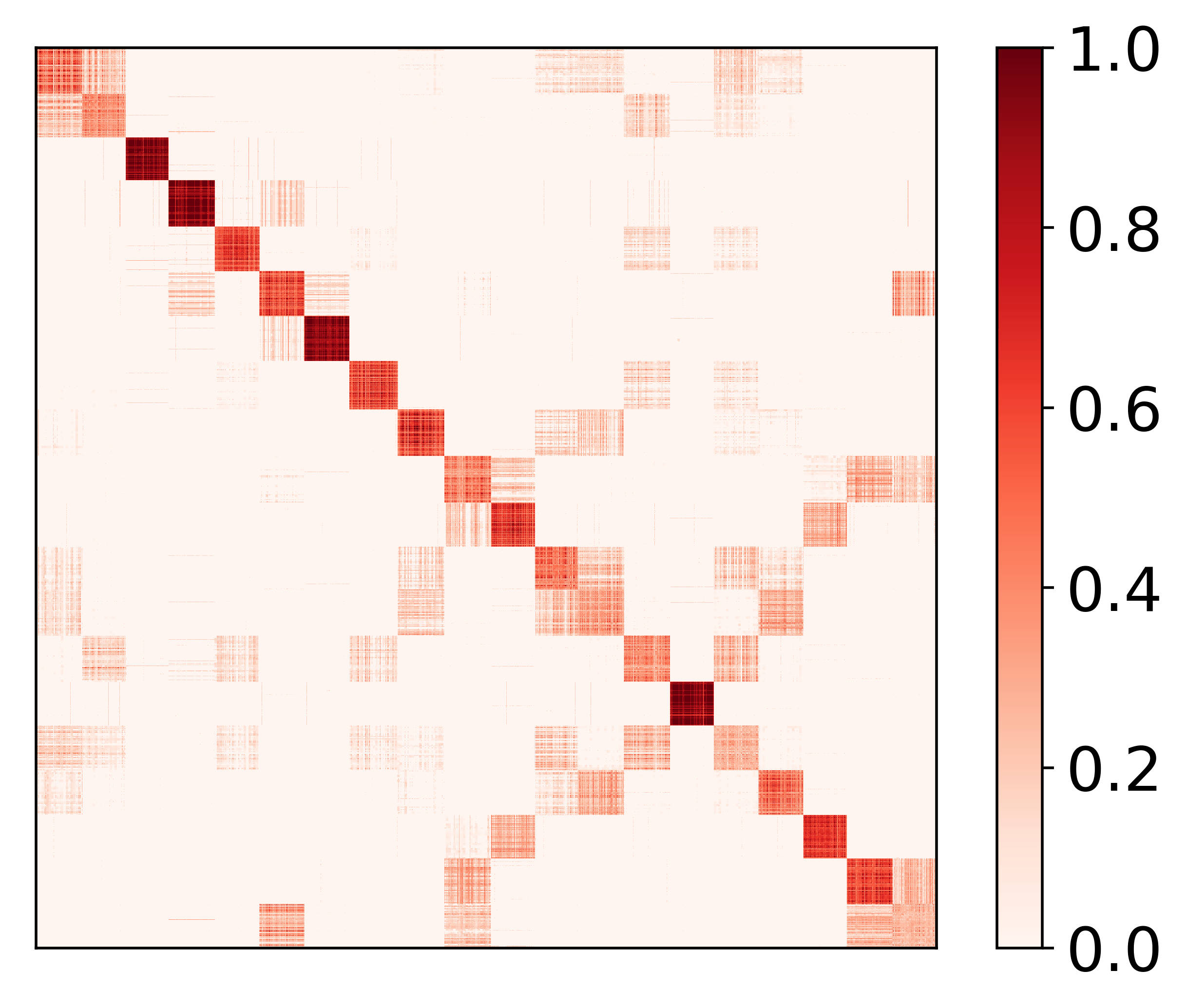}
        \caption{$\mathbf{\Pi}_\mathrm{\Phi}$ CIFAR-20}
    \end{subfigure}
    \begin{subfigure}[t]{0.32\linewidth}
        \includegraphics[width=\linewidth]{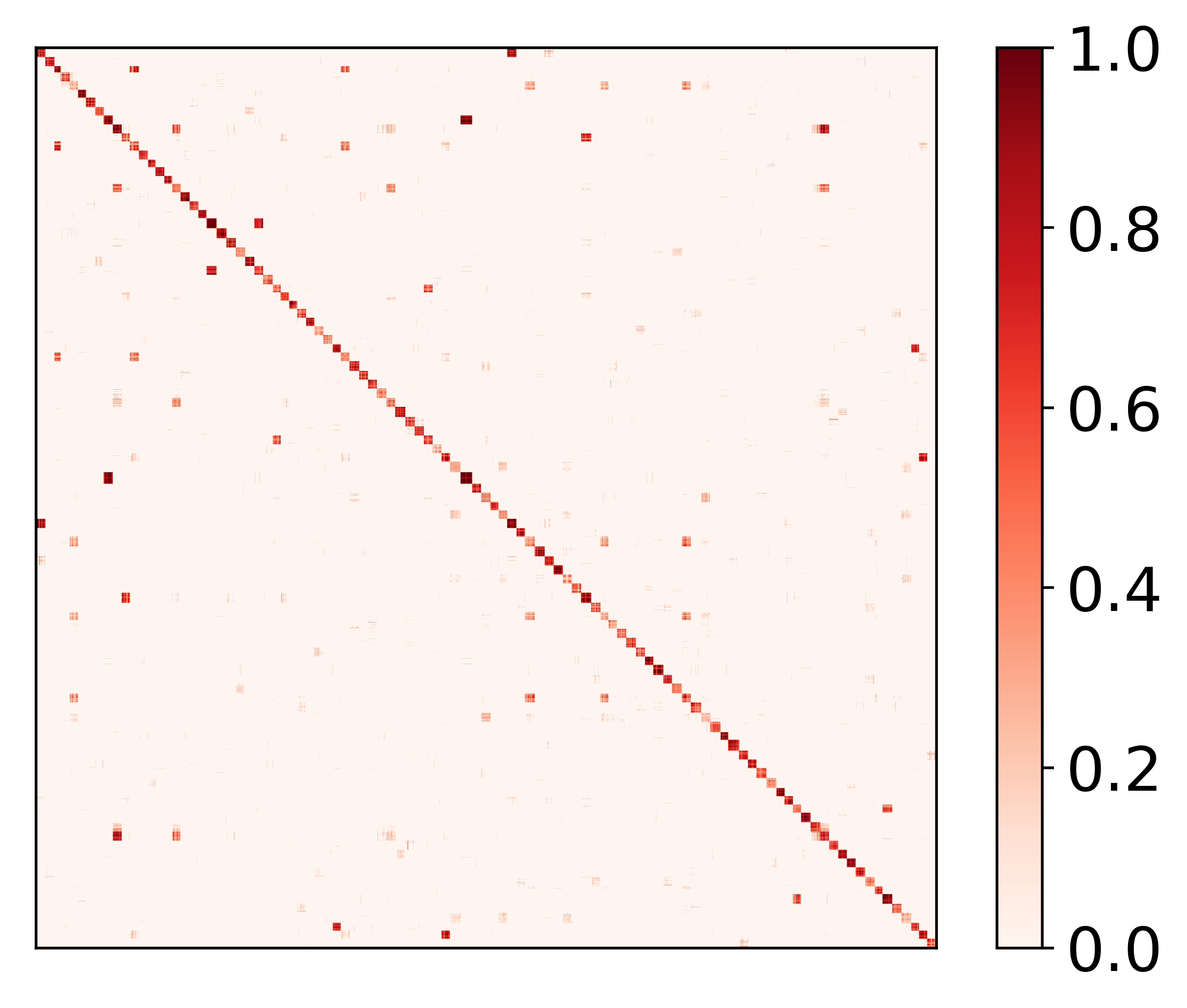}
        \caption{$\mathbf{\Pi}_\mathrm{\Phi}$ CIFAR-100}
    \end{subfigure}
    \begin{subfigure}[t]{0.32\linewidth}
        \includegraphics[width=\linewidth]{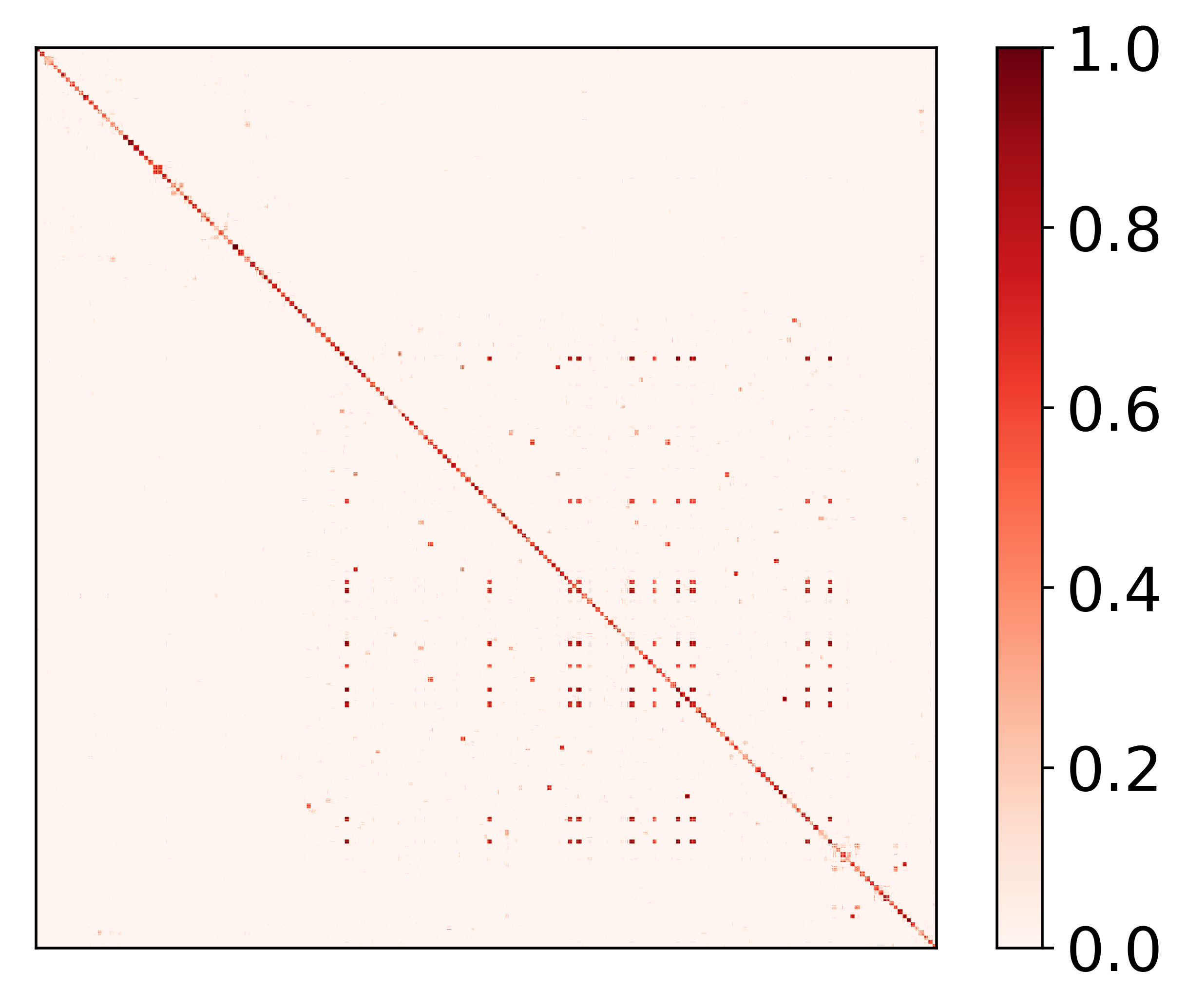}
        \caption{$\mathbf{\Pi}_\mathrm{\Phi}$ TinyImageNet}
    \end{subfigure}
\caption{\textbf{Ground-truth similarity matrices.} We plot the similarity matrices of \textbf{(a)-(c):} CLIP pre-features, \textbf{(d)-(f):} features produced by the CgMCR$^2$'s feature head, and \textbf{(g)-(i):} cluster memberships produced by the CgMCR$^2$'s cluster head on CIFAR-20, -100, and TinyImageNet, respectively.}
\label{fig:ground_truth}
\end{figure}

\myparagraph{Ground-truth similarity matrix}
The ground-truth similarity matrix is derived by computing by the cosine similarity  between data pairs belonging to the same ground-truth cluster.
An optimal ground-truth similarity matrix of representations or memberships should exhibit a block-diagonal structure aligned with the sorted ground-truth labels.
In Fig.~\ref{fig:ground_truth}, we visualize the ground-truth similarity matrices of CLIP pre-features, as well as the features and cluster memberships generated by CgMCR$^2$ on CIFAR-20, -100 and TinyImageNet.
The block-diagonal structures of the ground-truth similarity matrices in our method are clearer than that of the CLIP pre-features.

\subsection{Learning Curve}

\begin{figure}[t]
    \centering
    \begin{subfigure}[t]{0.3\linewidth}
        \includegraphics[width=0.95\textwidth]{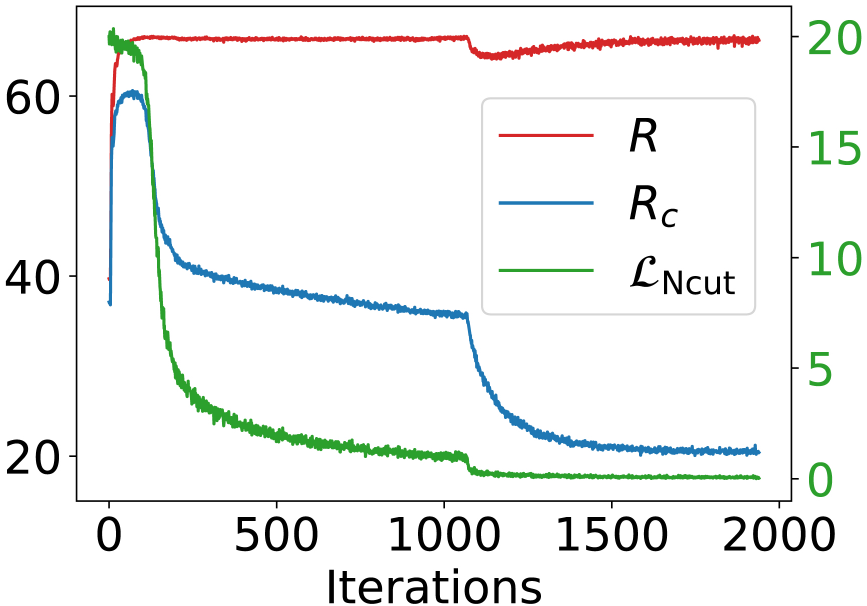}
        \caption{CIFAR-10}
    \end{subfigure}
    \begin{subfigure}[t]{0.3\linewidth}
        \includegraphics[width=0.95\textwidth]{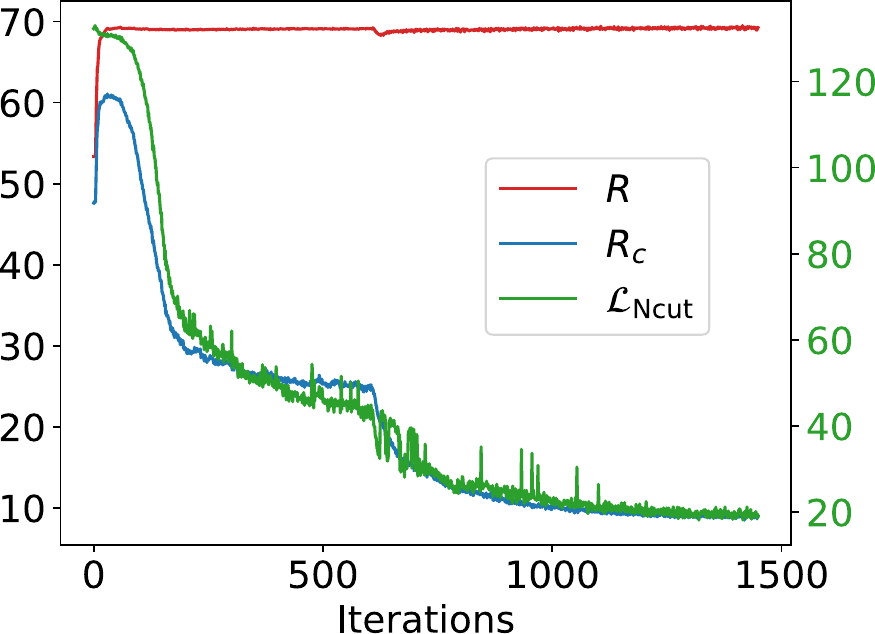}
        \caption{CIFAR-100}
    \end{subfigure}
    \begin{subfigure}[t]{0.3\linewidth}
        \includegraphics[width=0.95\textwidth]{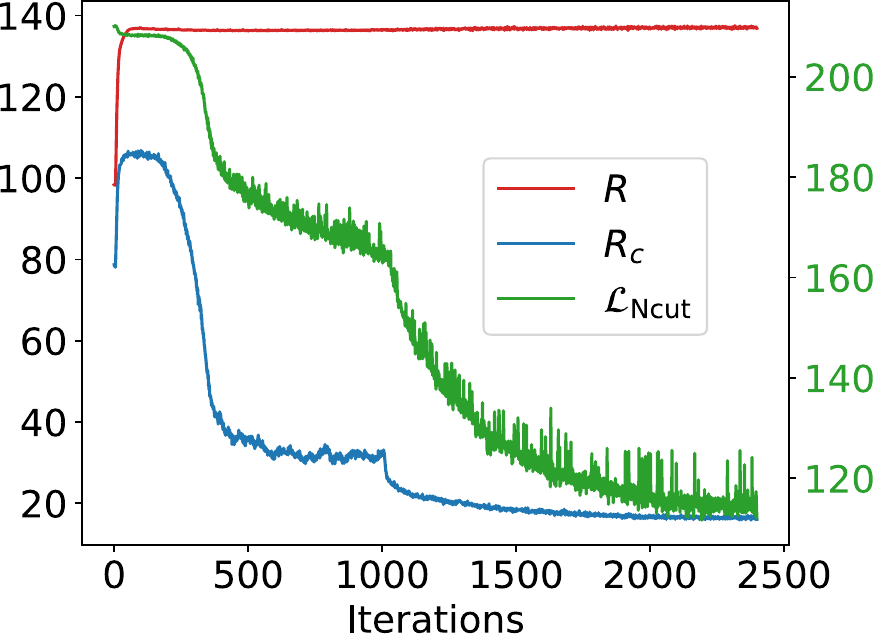}
        \caption{TinyImageNet}
    \end{subfigure} \\
    \begin{subfigure}[t]{0.3\linewidth}
        \includegraphics[width=0.95\textwidth]{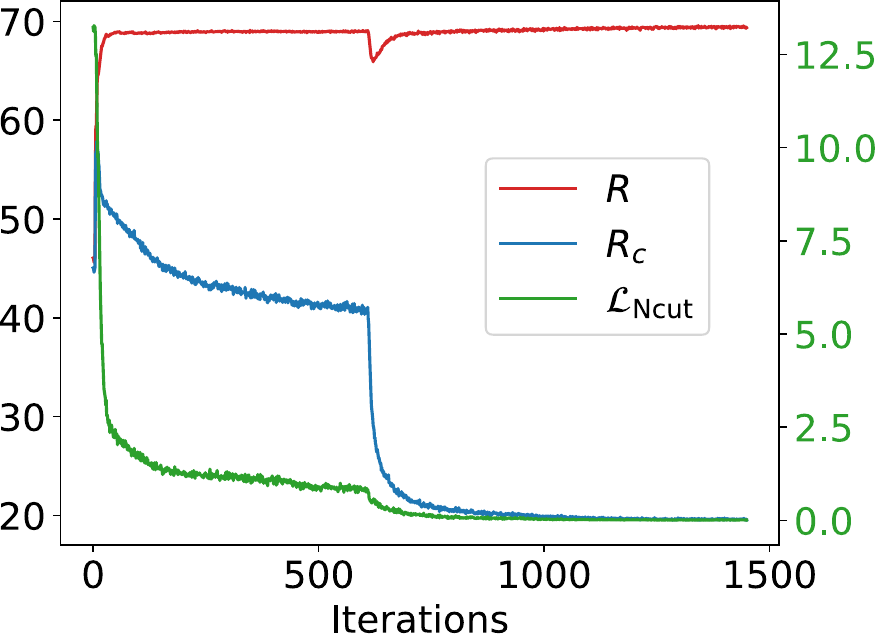}
        \caption{MNIST}
    \end{subfigure}
    \begin{subfigure}[t]{0.3\linewidth}
        \includegraphics[width=0.95\textwidth]{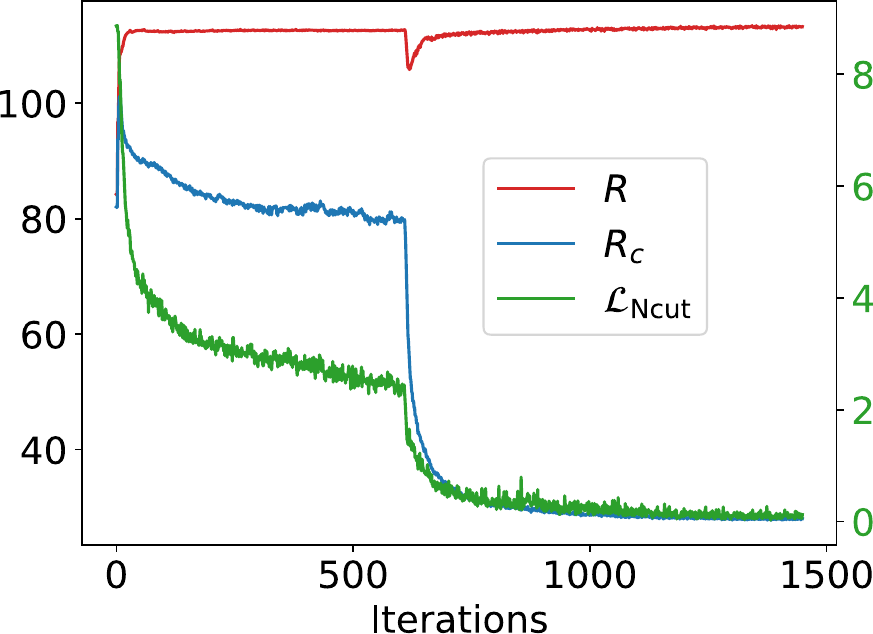}
        \caption{F-MNIST}
    \end{subfigure}
    \begin{subfigure}[t]{0.3\linewidth}
        \includegraphics[width=0.95\textwidth]{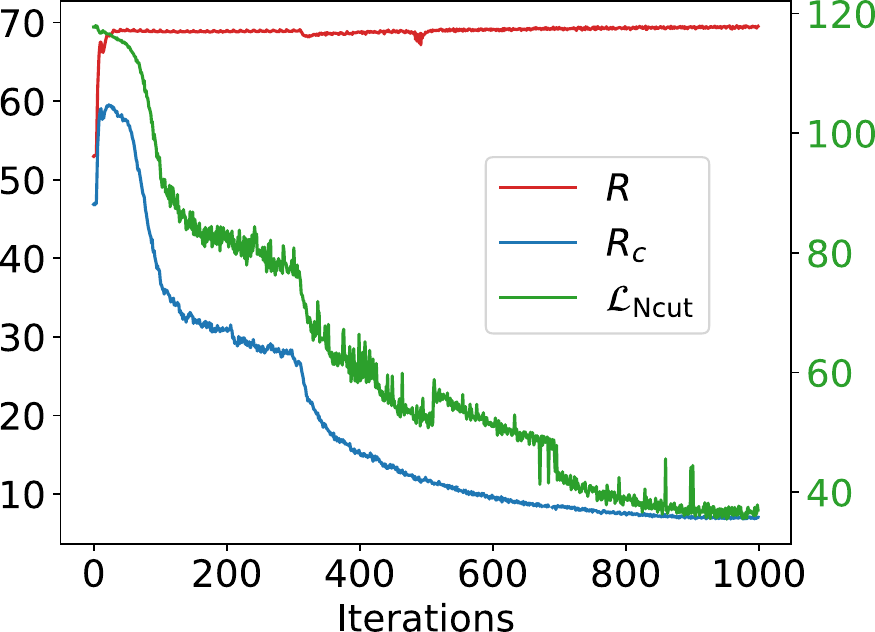}
        \caption{Dogs-120}
    \end{subfigure}
\caption{\textbf{Learning curves} of each term in the CgMCR$^2$ objective on CIFAR-10, -100, TinyImageNet, MNIST, F-MNIST and Dogs-120.}
\label{fig:more_obj_curve}
\end{figure}

\myparagraph{Loss curves}
In Fig.~\ref{fig:more_obj_curve}, we plot the loss curves of the CgMCR$^2$ objective during the training on CIFAR-10, -100, TinyImageNet, MNIST, F-MNIST and Dogs-120.
As can be seen, the variation of these loss terms are consistent across all datasets.
During the one-shot initialization, the term $R(\Z_\mathrm{\Theta};\epsilon)$ initially increases to its maximum to learn discriminative representations, and subsequently the term $\L_\mathrm{Ncut}(\mathbf{\Pi}_\mathrm{\Phi};\mathbf{A},\gamma)$ decrease to their \textit{local} minimum as it learns partition information from the discriminative representations.
During the fine-tuning, both $\L_\mathrm{Ncut}(\mathbf{\Pi}_\mathrm{\Phi};\mathbf{A},\gamma)$ and $R_c (\Z_\mathrm{\Theta}, {\mathbf \Pi}_\mathrm{\Phi};\epsilon)$ decrease to their global minimum, while the value of $R(\Z_\mathrm{\Theta};\epsilon)$ remains relatively constant.

\begin{figure}[!t]
    \centering
    \begin{subfigure}[t]{0.3\linewidth}
        \includegraphics[width=0.95\textwidth]{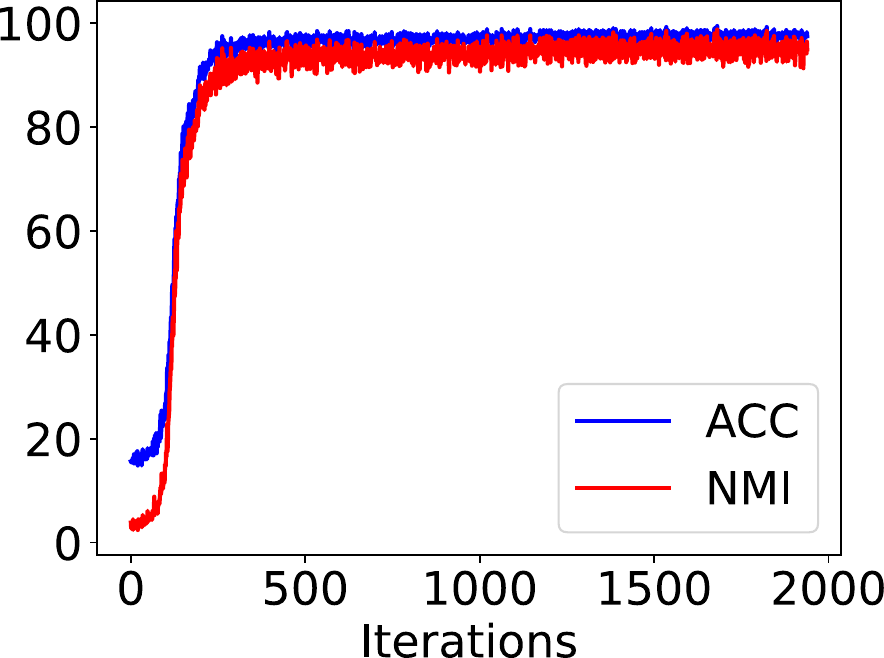}
        \caption{CIFAR-10}
    \end{subfigure}
    \begin{subfigure}[t]{0.3\linewidth}
        \includegraphics[width=0.95\textwidth]{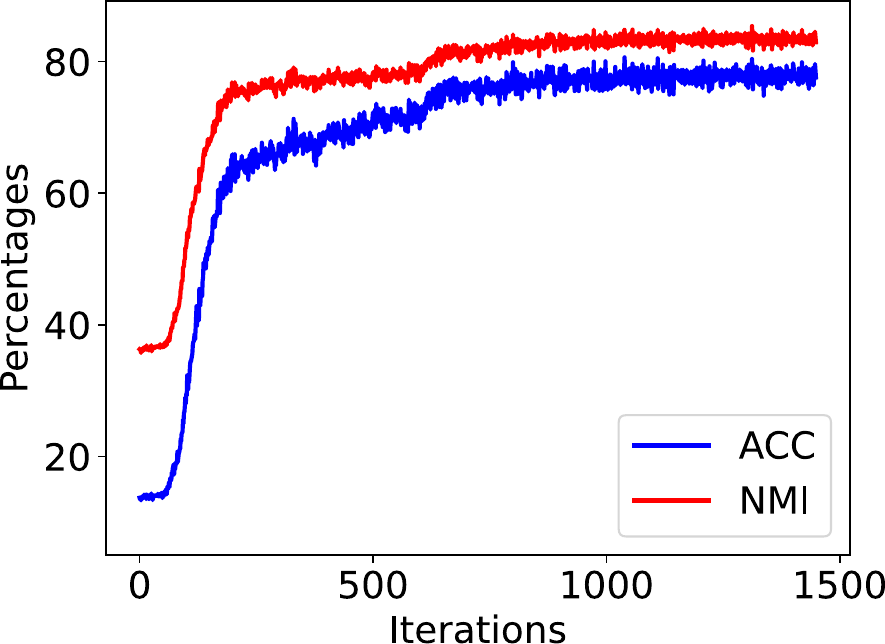}
        \caption{CIFAR-100}
    \end{subfigure}
    \begin{subfigure}[t]{0.3\linewidth}
        \includegraphics[width=0.95\textwidth]{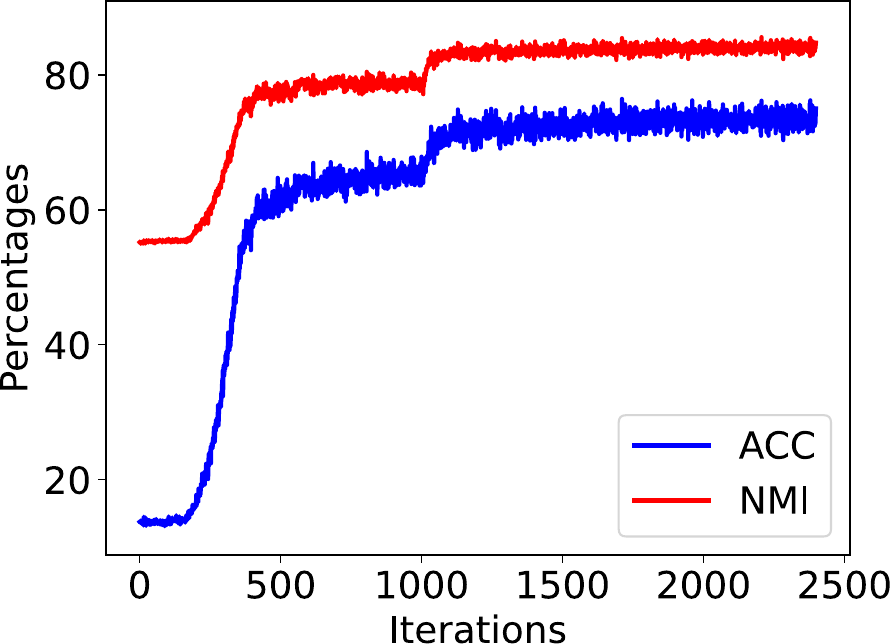}
        \caption{TinyImageNet}
    \end{subfigure} \\
    \begin{subfigure}[t]{0.3\linewidth}
        \includegraphics[width=0.95\textwidth]{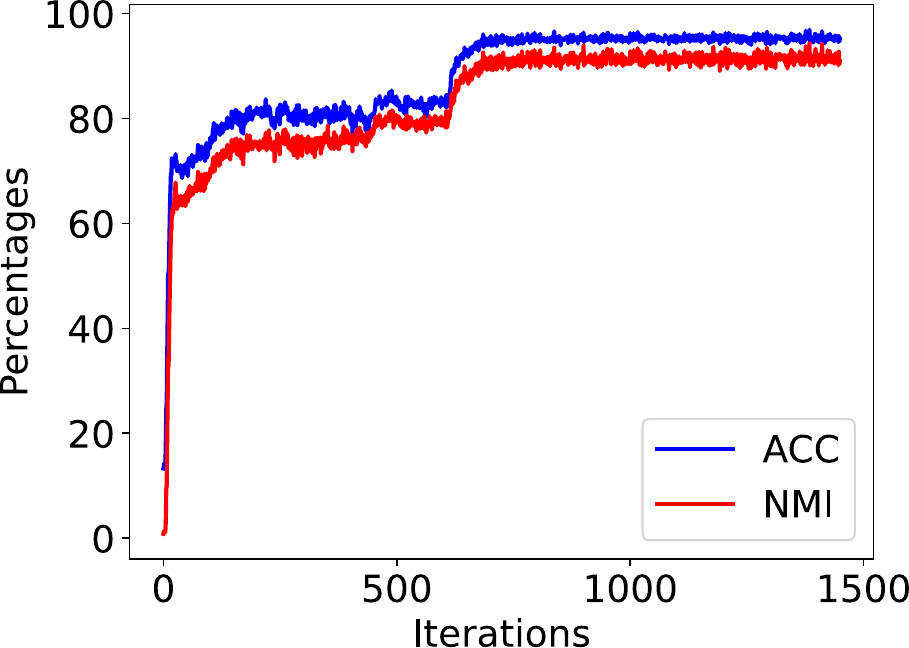}
        \caption{MNIST}
    \end{subfigure}
    \begin{subfigure}[t]{0.3\linewidth}
        \includegraphics[width=0.95\textwidth]{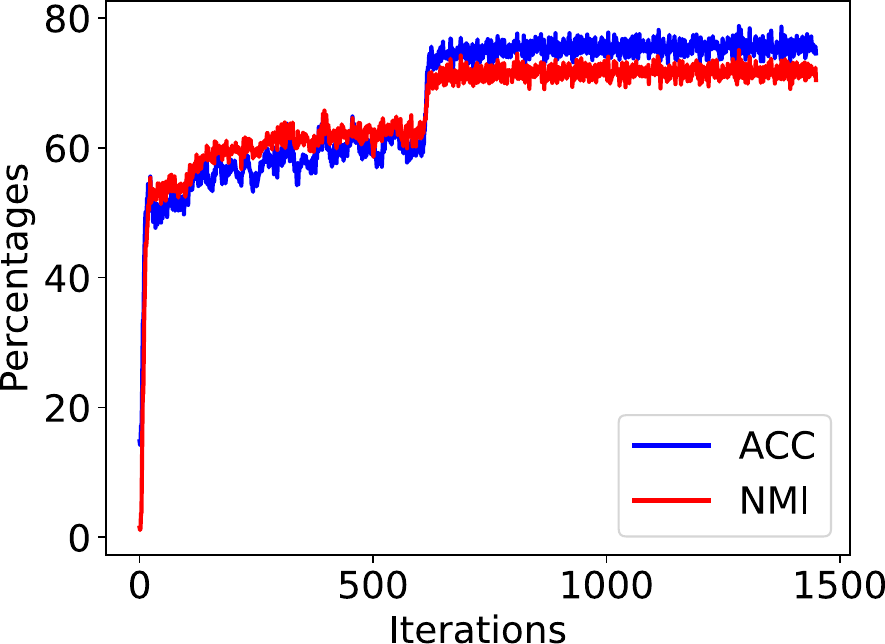}
        \caption{F-MNIST}
    \end{subfigure}
    \begin{subfigure}[t]{0.3\linewidth}
        \includegraphics[width=0.95\textwidth]{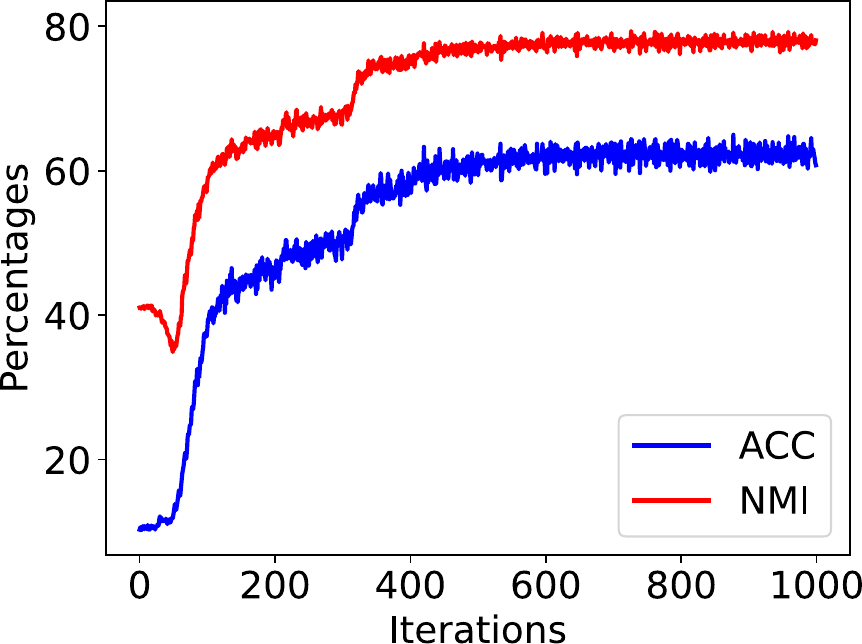}
        \caption{Dogs-120}
    \end{subfigure}
\caption{\textbf{ACC and NMI curves} of $\mathbf{\Pi}_\mathrm{\Phi}$ on CIFAR-10, -100, TinyImageNet, MNIST, F-MNIST and Dogs-120.}
\label{fig:more_acc_curve}
\end{figure}

\myparagraph{ACC and NMI curves}
We take the outputs of the cluster head $\mathbf{\Pi}_\Theta$ as the cluster membership and plot its ACC and NMI during each training iteration on CIFAR-20, -100, TinyImageNet, MNIST, F-MNIST and Dogs-120 datasets.
In Fig.~\ref{fig:more_acc_curve}, we can observe that our CgMCR$^2$ converges and achieves the stable clustering results on all tested datasets within 2,500 training iterations.

\end{document}